\documentclass[a4paper,fleqn]{cas-dc}

\usepackage[numbers,sort&compress]{natbib}
\usepackage{tabularx}
\usepackage{bbding}
\usepackage{makecell}
\usepackage{array}
\usepackage{multirow}
\usepackage{enumitem}
\usepackage{algorithm}  
\usepackage{algorithmic}
\usepackage{hyperref}
\hypersetup{
    colorlinks=true,
    linkcolor=blue,
    urlcolor=blue,
    anchorcolor=blue,
    citecolor=blue
}

\usepackage{float}
\usepackage{tikz}
\usetikzlibrary{backgrounds,mindmap}
\usepackage{pgfplots}
\usepackage{listings}
\usepackage{soul}
\usepackage{color, xcolor}  
\sethlcolor{yellow}

\begin{document}

\shorttitle{Global–focal Adaptation with Information Separation for Noise‑robust Transfer Fault Diagnosis} 
\shortauthors{Junyu Ren \textit{et al.}}

\title [mode = title]{Global–focal Adaptation with Information Separation for Noise‑robust Transfer Fault Diagnosis}  

\author[1]{Junyu Ren}
\ead{renjunyu193@gmail.com}
\address[1]{Guangdong Lingnan Institute of Technology, Qingyuan 511500, China}

\author[2]{Wensheng Gan}
\ead{wsgan001@gmail.com}
\address[2]{Jinan University, Guangzhou 510632, China}
\cortext[cor1]{Corresponding author}
\cormark[1]

\author[1]{Guangyu Zhang}
\ead{TchrZgy7626@163.com}

\author[1]{Wei Zhong}
\ead{chungchung718991776@gmail.com}

\author[3]{Philip S. Yu}
\ead{psyu@uic.edu}
\address[3]{University of Illinois Chicago, Chicago 60607, USA} 

\begin{abstract}
    Existing transfer fault diagnosis methods typically assume either clean data or sufficient domain similarity, which limits their effectiveness in industrial environments where severe noise interference and domain shifts coexist. To address this challenge, we propose an information separation global-focal adversarial network (ISGFAN), a robust framework for cross-domain fault diagnosis under noise conditions. ISGFAN is built on an information separation architecture that integrates adversarial learning with an improved orthogonal loss to decouple domain-invariant fault representation, thereby isolating noise interference and domain-specific characteristics. To further strengthen transfer robustness, ISGFAN employs a global-focal domain-adversarial scheme that constrains both the conditional and marginal distributions of the model. Specifically, the focal domain-adversarial component mitigates category-specific transfer obstacles caused by noise in unsupervised scenarios, while the global domain classifier ensures alignment of the overall distribution. Experiments conducted on three public benchmark datasets demonstrate that the proposed method outperforms other prominent existing approaches, confirming the superiority of the ISGFAN framework. Data and code are available at \url{https://github.com/JYREN-Source/ISGFAN}
\end{abstract}


\maketitle

\section{Introduction} \label{sec:introduction}

Rotating machinery \citep{shi2025bearing} is critical in industrial applications, where system reliability is essential to avoid financial losses and safety risks. Therefore, timely fault diagnosis is a crucial engineering priority. Deep learning-based fault diagnosis has achieved remarkable success due to its ability to extract features and model complex nonlinear relationships \citep{ li2025novel,zhu2023review}. However, industrial rotating machines operate under diverse conditions, leading to domain shifts that degrade the diagnostic performance of conventional deep learning methods \citep{misbah2024fault}. Among the powerful artificial intelligence (AI) technologies, transfer learning \citep{pan2009survey} can address these limitations through cross-task knowledge transfer, where domain adaptation has become a widely adopted technique in fault diagnosis, primarily encompassing metric-based approaches, adversarial frameworks, and their hybrid variants \citep{misbah2024fault, chen2023deep}. Currently, cross-domain fault diagnosis methods have been extended to encompass a wider range of diverse and practical application scenarios \citep{chen2025review}. Given that source domain data are often more abundant in real-world settings, several studies have proposed multi-source transfer fault diagnosis approaches \citep{tang2024parallel, cui2025two}. For closed-set scenarios, various domain adaptation methods have been developed \citep{qian2024new}. Since the label categories between source and target domains may not be completely identical, open-set domain adaptation and partial domain adaptation methods have been developed for fault diagnosis \citep{su2023multi}. Universal domain adaptation offers a unified framework for addressing label mismatch issues in fault diagnosis \citep{liu2025universal}. Additionally, some studies have considered scenarios where small samples and domain shift occur simultaneously \citep{wang2025progressive}. Due to data privacy requirements across different devices, source-free domain adaptation methods for fault diagnosis have been developed \citep{gao2024multi}. However, most existing transfer fault diagnosis methods assume clean data, neglecting noise interference in industrial environments, which substantially complicates the transfer process.

The intricate operational environments of rotating machinery often induce noise interference from both external and internal sources, characterized by randomness and persistence (e.g., stochastic vibrations from rough road surfaces, sustained noise from other industrial equipment, and inherent structural vibrations within the machinery). Some studies have increasingly explored noise-robust intelligent fault diagnosis methods, focusing primarily on enhancing feature extraction from multiple perspectives and leveraging inter-sample correlations to achieve more accurate decision boundaries \citep{guo2024attention}. Representative techniques include multi-scale feature extractors \citep{gao2024multi1}, improved attention mechanisms \citep{han2024amcw}, heterogeneous convolutional operators \citep{huang2024deep}, and graph convolutional networks (GCN) \citep{he2025agfcn}, which collectively improve diagnostic accuracy in noisy environments. While these approaches effectively address noise-resistant fault diagnosis, they overlook model generalization and distribution shift issues in cross-domain scenarios.

\begin{figure}
  \centering
  \includegraphics[width=0.95\linewidth]{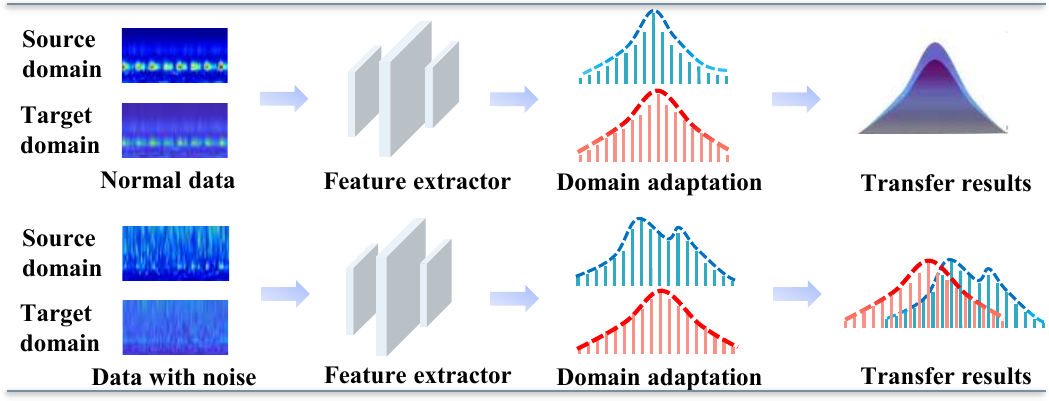}
  \caption{Transfer processes across various situations. }
  \label{fig1}
\end{figure}

Cross-domain fault diagnosis \citep{chen2023transfer} and noise-robust fault diagnosis \citep{zhu2023review} are typically investigated separately. However, industrial environments often involve both noise and domain shifts. Noise-robust fault diagnosis methods often suffer from limited generalization performance, while most existing transfer fault diagnosis methods primarily focus on directly aligning domain distributions and assume clean data. Their effectiveness largely depends on sufficient similarity between the source and target domains \citep{wang2022generalizing}. However, severe noise interference introduces substantial irrelevant information while masking domain-invariant fault discriminative features, inadvertently exacerbating domain discrepancies that become difficult to estimate. Furthermore, noise of equivalent intensity may differentially obscure fault discriminative features across different categories, resulting in transfer difficulties for specific classes. Under such conditions, direct modeling of global or local distribution consistency risks inducing erroneous fitting to fault-irrelevant information, thereby diminishing the salience of domain-invariant representations, misdirecting the optimization of decision boundaries, and substantially compromising transfer efficacy. As illustrated in Figure \ref{fig1}, existing fault diagnosis methods struggle to achieve excellent performance due to excessive domain discrepancies caused by noise interference.

To address these challenges, we propose an information separation global-focal adversarial network for cross-domain fault diagnosis under noise interference. First, to mitigate noise interference during transfer, we design an information separation framework that constructs an auxiliary guidance branch based on adversarial training to model fault-irrelevant information—such as noise and domain-private features—while maximizing the discrepancy between fault-relevant and fault-irrelevant features through an improved orthogonality loss. This enables the model to focus on domain-invariant fault representations. Building upon these purified domain-invariant representations, a global-focal domain adversarial scheme is proposed for comprehensive distribution alignment. Global domain adaptation is used for marginal distribution alignment, while the focal domain adaptation component utilizes a subdomain adversarial strategy and a subdomain attention algorithm to address noise-induced category-specific transfer obstacles, constraining conditional distributions adaptively. Subsequently, a dynamic loss-weighting strategy is employed to counteract gradient dominance in the multi-task learning process. The main innovations and contributions of this paper can be summarized as follows:

\begin{itemize}
    \item This study explores the impact of noise on transfer fault diagnosis. The proposed ISGFAN algorithm addresses this research gap by enabling robust knowledge transfer under noisy conditions.
    
    \item An information separation framework is developed to decouple noise and domain-specific features from complex data, yielding cleaner domain-invariant fault representations for domain adaptation and classification tasks.
    \item A global-focal domain adversarial scheme is proposed, which, under noise interference,  effectively aligns marginal and conditional distributions to achieve comprehensive adversarial domain adaptation.
    \item In the presence of substantial noise interference, ISGFAN achieves average cross-domain accuracies of 88.53\%, 85.03\%, and 78.39\% across multiple transfer tasks in three datasets, surpassing all comparison models.
\end{itemize}

The remainder of this paper is organized as follows: Section \ref{sec: Related Work} introduces the related works. Section \ref{sec:preliminaries} introduces the theoretical background and preliminaries. Section \ref{sec:Proposed method} presents the proposed methodology in detail. Section \ref{sec: Experiment and Analysis} presents the experiments and results analysis. Finally, Section \ref{sec:Conclusion} summarizes the contributions of this paper and outlines future work.

\section{Related Work} \label{sec: Related Work}
In this section, we provide a brief review of recent intelligent fault diagnosis methods related to this study, which can be categorized into noise-robust fault diagnosis and transfer fault diagnosis.

\subsection{Noise-robust Fault Diagnosis}

Noise-resistant fault diagnosis aims to mitigate noise interference through two main strategies: (1) bolstering feature extraction mechanisms to yield robust representations, exemplified by techniques such as multi-scale convolutions, attention modules, and time-frequency transforms; and (2) refining decision boundaries for enhanced separability, as achieved through methods like contrastive learning, graph structures, and clustering algorithms. To date, numerous studies have advanced this field. \citet{guo2024attention} proposed an attention-enhanced ConvNeXt for adaptive feature extraction and denoising, while \citet{wang2025making} introduced a multi-level supervised framework for noisy environments. Addressing CNN limitations, \citet{huang2024deep} developed a deep continuous convolutional network, and \citet{gao2024multi1} created an adaptive multi-timescale attention network for noise-robust fault diagnosis. \citet{he2025agfcn} advanced graph-based approaches with the Adaptive Graph Framelet Convolutional Network to exploit sample correlations. Further, \citet{chen2025noise} designed a noise-resilient residual network, \citet{fan2025motor} proposed the LMSWT-SE-MSCNN for fine-grained feature extraction, \citet{han2024amcw} integrated wavelet transforms in a deep feature fusion network for adaptive denoising, and \citet{li2025graph} optimized graph structures via dual-scale spectral features and contrastive learning for noisy, low-label settings.

These noise-robust methods have advanced intelligent fault diagnosis toward practical applications, yet inherent limitations remain. In severe noise, obscured inter-sample correlations weaken decision boundary refinement strategies, while feature extraction enhancements often require substantial computational resources. Additionally, by prioritizing accuracy over generalization, these approaches falter when domain shifts coincide with noise, limiting broader applicability.

\subsection{Transfer Fault Diagnosis}
Transfer fault diagnosis aims to mitigate domain discrepancy \citep{misbah2024fault}. Mainstream methods include metric-based and adversarial-based approaches, along with variants integrating multiple techniques. The core objective remains learning domain-invariant universal representations \citep{yang2025enhancing, zhang2025application, jia2025physics}. \citet{gao2024multi} introduced a source-free domain transfer method for privacy-preserving fault diagnosis. \citet{qian2024new} proposed a Gaussian distribution-guided indirect domain adaptation scheme for scenarios with large domain gaps. \citet{tang2024parallel} created a parallel ensemble optimization loss function with a multi-source transfer learning framework. To address category shifts in open-set fault diagnosis, \citet{su2023multi} proposed a multi-adversarial deep transfer network with fault class matching. \citet{chen2023collaborative} developed a dual adversarial guided network combining edge and inner adversarial modules for enhanced multi-domain adaptation. \citet{yang2024novel} introduced a dual-network autoencoder with adversarial domain adaptation using Wasserstein divergence. \citet{shao2024adaptive} built an adaptive multi-scale attention CNN for cross-domain detection with limited data. \citet{an2023domain} presented a contrastive learning-based domain adaptation network to mitigate boundary misclassification. To tackle faults with domain shifts and class imbalance, \citet{lee2024domain} developed a domain adversarial learning framework with label-aligned sampling. \citet{zhang2025fed} proposed a federated learning framework for transfer fault diagnosis, integrating interpretable wavelet fusion networks and pruning training. 

Transfer fault diagnosis significantly enhances model generalization but retains limitations. Current methods directly align global or local domain distributions, relying on assumptions of (1) sufficient inter-domain similarity and (2) equal transfer difficulty across subdomains. These overlook industrial noise, which introduces irrelevant information, obscures discriminative features, and indirectly exacerbates domain gaps. Moreover, noise widens subdomain transfer disparities, impeding optimization for poorly transferring categories when subdomains are treated equally. Drawing on limitations in noise-robust and transfer fault diagnosis, we address a more practical industrial scenario: transfer diagnosis under strong noise. Our ISGFAN decouples information instead of forcing alignment, incorporating a global-focal adaptation method that unsupervisedly tackles high-difficulty subdomains while ensuring global distribution alignment.

\section{Preliminaries}  \label{sec:preliminaries}
\subsection{Problem Setting}

In cross-domain fault diagnosis, knowledge from one operating condition (source domain) is transferred to another condition (target domain) \citep{pan2009survey}. Formally, we have a source domain $\mathcal{D}_s$ = $\{(x_i^s, y_i^s)\}_{i=1}^{n_s}$ with $n_s$ labeled examples where $y_i^s$ $\in \{1, 2, \ldots, C\}$, and a target domain $\mathcal{D}_t$ = $\{x_i^t\}_{i=1}^{n_t}$ with $n_t$ unlabeled examples. The source and target domains are drawn from different probability distributions $P$ and $Q$ where $P \neq Q$. The objective is to design a deep neural network $y$ = $f(x)$ that reduces cross-domain distribution shifts and learns transferable representations.
 
\subsection{Adversarial Domain Adaptation}

\begin{figure}
    \centering
    \includegraphics[width=1\linewidth]{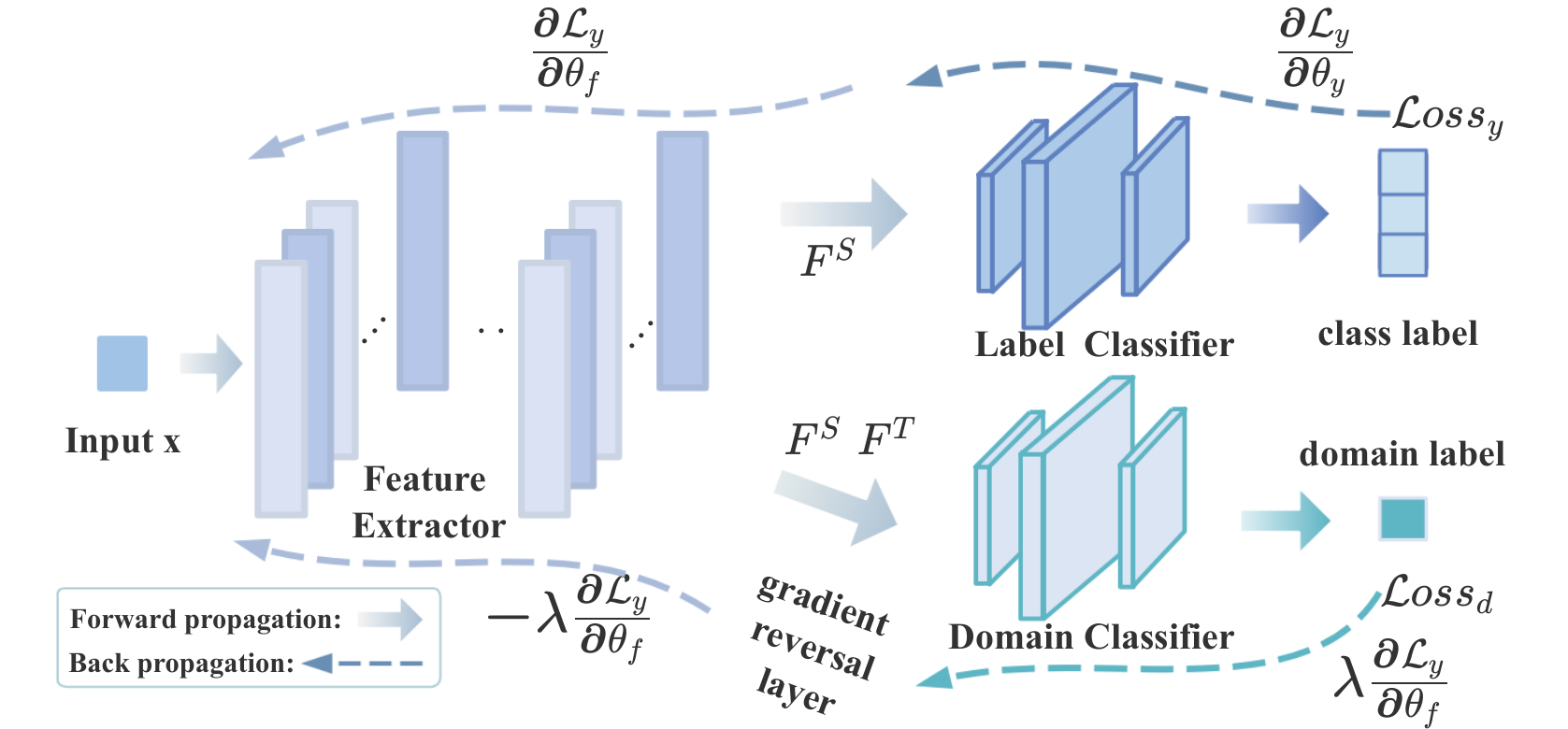}
    \caption{The framework of DANN. }\label{basedDANN}
\end{figure}

Inspired by Generative Adversarial Networks (GANs) \citep{goodfellow2020generative}, adversarial-based domain adaptation methods have been extensively developed. Among them, the classical Domain-adversarial neural Networks (DANN) \citep{ganin2016domain} employ adversarial training to learn robust, domain-invariant feature representations. 

The detailed framework of DANN is illustrated in Figure \ref{basedDANN}. DANN introduces a Gradient Reversal Layer (GRL) as its key component, which acts as an identity transform during forward propagation but multiplies gradients by a negative scalar during backpropagation. The behavior of $\mathcal{R}(x)$ is defined by the following equations, which describe its forward and backpropagation characteristics:
\begin{align}
    \mathcal{R}(x) & = x, \label{eq:16} \\
    \frac{d\mathcal{R}}{dx} & = -\mathbf{I}. \label{eq:17}
\end{align}
where $\mathbf{I}$ is an identity matrix. This mechanism compels the feature extractor to learn discriminative features for the main classification task and be invariant to domain shifts. The training process can be summarized as optimizing the min-max objective described above:

\begin{equation}
    \label{eq:dann-obj}
    \min_{\theta_f, \theta_y} \max_{\theta_d} \Bigl[ \mathcal{L}_y(\theta_f, \theta_y) - \lambda \mathcal{L}_d(\theta_f, \theta_d) \Bigr],
\end{equation}
where $\mathcal{L}_y(\theta_f, \theta_y)$ is the classification loss for predicting source domain labels, and $\mathcal{L}_d(\theta_f, \theta_d)$ is the domain classification loss for distinguishing between source and target domains.

\section{Proposed Method}  \label{sec:Proposed method}

\begin{figure}
    \centering
    \includegraphics[width=1\linewidth]{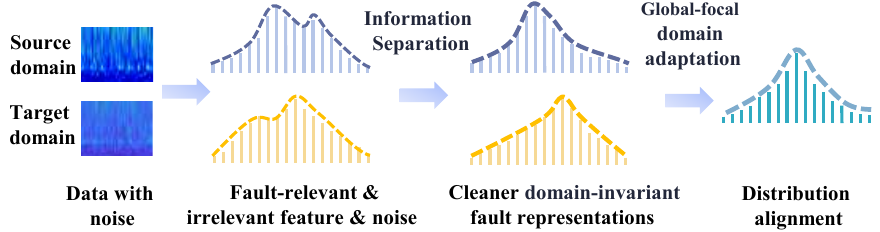}
    \caption{Conceptual illustration of the proposed method.}
    \label{Conceptual illustration of the proposed method}
\end{figure}

\begin{figure*}
    \centering
    \includegraphics[width=0.90\linewidth]{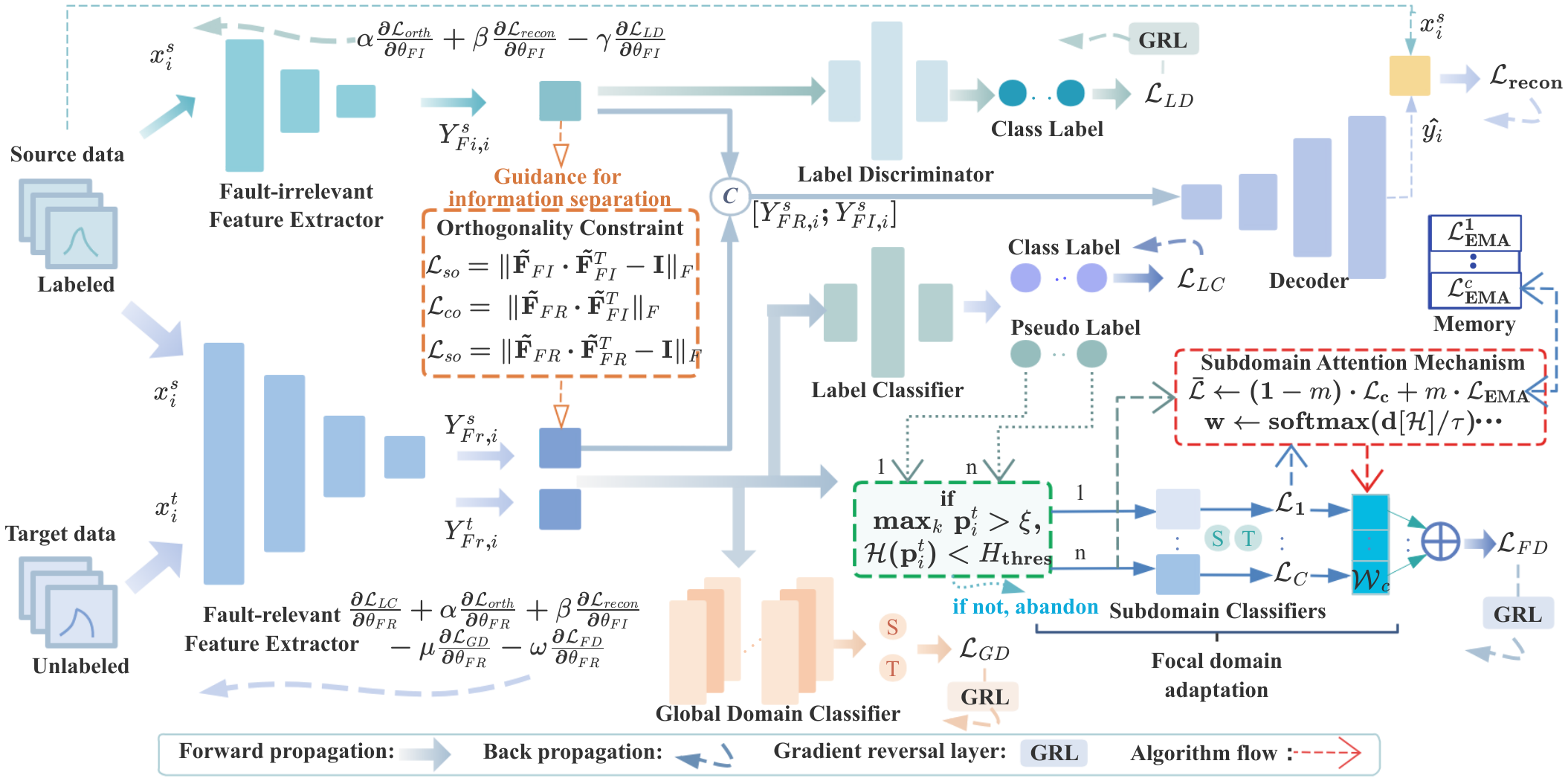}
    \caption{The framework of information separation global-focal adversarial network.}
    \label{Fig-ISGFAN}
\end{figure*}

Overall, ISGFAN is an end-to-end diagnostic model that uses one-dimensional vibration signals, with its architecture illustrated in Figure \ref{Fig-ISGFAN}. ISGFAN operates through two key components: an information separation framework and a global-focal domain adversarial module. Figure \ref{Conceptual illustration of the proposed method} provides a conceptual illustration of the function of these two components. The separation framework comprises the fault-relevant feature extractor (FRFE), fault-irrelevant feature extractor (FIFE), label discriminator (LD), Decoder, and label classifier (LC). The domain adversarial module leverages the FRFE, global domain classifier (GDC), and subdomain classifiers (SDC). The FRFE and LC together form the primary branch for testing.

\subsection{Information Separation Framework} 

Under noisy conditions, the vibration signals are primarily composed of three types of information: domain-invariant fault-relevant representations, domain-specific features, and noise. The latter two can be categorized as fault-irrelevant information, which harms the transfer process. Inspired by the theories of disentangled representation learning \cite{wang2024disentangled} and domain-adversarial training \cite{ganin2016domain}, an information separation framework has been proposed. The FIFE and LD form an information separation guiding branch to extract fault-irrelevant information, while an improved orthogonal constraint is introduced to help the main branch focus on domain-invariant fault representations. This approach mitigates the impact of noise and source domain-specific information on diagnostic accuracy and the transfer process.

Specifically, a pair of feature extractors has been constructed. The FRFE operates simultaneously on both domains for extracting domain-invariant features, while the FIFE functions independently on the source domain to model information represented by noise and domain-specific features. This process can be expressed as follows:
\begin{align}  
Y_{FR,i}^{s,t} &= E_{FR}\left(x_i^{s,t}; \theta_{FR}\right) \label{eq7} \\[6pt]  
Y_{FI,i}^{s} &= E_{FI}^{s}\left(x_i^{s}; \theta_{FI}\right) \label{eq8} 
\end{align}
where \(Y_{FR,i}^{s,t}\), \(E_{FR}\) and \(\theta_{FR}\) are the output, function, and parameters of FRFE for the \(i\)-th sample \(x_i^{s,t}\) from either the source or target domain. Likewise, \(Y_{FI,i}^{s}\), \(E_{FI}\), and \(\theta_{FI}\) are the output, function, and parameters of FIFE.

The high-dimensional features \(Y_{FR,i}^{s}\) and \(Y_{FI,i}^{s}\) are fed into LC and LD, respectively. Each module \(m \in \{\mathrm{LC}, \mathrm{LD}\}\) employs cross-entropy \citep{rumelhart1986learning} as its loss function, defined as follows:
\begin{align}  
\mathcal{L}^{(m)} = - \frac{1}{n_s} \sum_{i=1}^{n_s} \sum_{j=1}^{C} y^{(m)}_{i,j} \log \bigl(\hat{y}^{(m)}_{i,j}\bigr).   
\label{eq:cross_entropy}  
\end{align}   
where \(\hat{y}^{(m)}_{i,j}\) is the predicted probability of sample \(i\) belonging to class \(j\), and \(y^{(m)}_{i,j}\) indicates the ground-truth class in one-hot encoding.

FRFE and LC update through standard backpropagation, while the GRL connecting FIFE and LD forces FIFE to learn fault-irrelevant features. Their adversarial interaction is expressed as follows:
\begin{equation}  
\frac{\partial \mathcal{L}_{LD}}{\partial \theta_{FI}}
= -\lambda \frac{\partial \mathcal{L}_{LD}}{\partial \theta_{LD}}
\label{eq:grl_impact}  
\end{equation}

Building on the orthogonal constraint \citep{liu2025universal}, an improved orthogonality loss incorporating cross-orthogonality and self-orthogonality terms was developed to guide the FRFE toward domain-invariant fault representations. The cross-term disentangles the two types of information by minimizing the Frobenius norm \citep{higham1988computing} of the correlation matrix between FIFE and FRFE outputs, ensuring their independence. The self-term independently drives the normalized feature Gram matrix within each feature extractor toward the identity matrix, reducing inter-channel correlations and eliminating redundant, task-irrelevant features. The synergy of these two terms yields purified and mutually disentangled feature representations from both extractors, enabling the FRFE to maximally separate noise and domain-specific information. The improved orthogonality loss is computed as:
\begin{equation}   
{\mathcal{L}_{co}} =  \|\tilde{\mathbf{F}}_{FR} \cdot \tilde{\mathbf{F}}_{FI}^T\|_F 
\label{eq:cortho} 
\end{equation}
\begin{equation}  
\mathcal{L}_{so} = \frac{1}{2}\left( \|\tilde{\mathbf{F}}_{FR} \cdot \tilde{\mathbf{F}}_{FR}^T - \mathbf{I}\|_F + \|\tilde{\mathbf{F}}_{FI} \cdot \tilde{\mathbf{F}}_{FI}^T - \mathbf{I}\|_F \right)
\label{eq:sortho}
\end{equation}
\begin{equation}  
\mathcal{L}_{orth} = \mathcal{L}_{co} + \mathcal{L}_{so}
\label{eq:ortho}
\end{equation}
where \(\|\cdot\|_F\) denotes the Frobenius norm. \(\mathbf{I}\) is the identity matrix.
$\tilde{\mathbf{F}}_{FR}$, $ \tilde{\mathbf{F}}_{FI}$ are the feature matrices from FRFE and FIFE after L2 normalization, respectively. $\mathcal{L}_{co}$, $\mathcal{L}_{so}$ represent the cross-orthogonality and self-orthogonality loss respectively.

To preserve complete information, concatenate \(Y_{FR,i}^{s}\) and \(Y_{FI,i}^{s}\) along the channel dimension to obtain \(Y_{concat,i}^{s}\), then pass it to the decoder:
\begin{align}
Y_{concat,i}^{s} &= [Y_{FR,i}^{s}; Y_{FI,i}^{s}] \\
\hat{y}_i &= \text{Decoder}\bigl(Y_{concat,i}^{s}; \theta_{dec}\bigr)
\end{align}
The decoder output \(\hat{y}_i\) is compared with the original input \(x_i^{s}\) through mean-squared error \citep{vincent2010stacked}:
\begin{align}
\mathcal{L}_{\text{mse}}(x_i^{s}, \hat{y}_i) &= \frac{1}{L}\,\|x_i^{s} - \hat{y}_i\|_2^{2} \\
\mathcal{L}_{\text{recon}} &= \sum_{i=1}^{n_s} \mathcal{L}_{\text{mse}}(x_i^{s}, \hat{y}_i)
\end{align}
where \(n_s\) is the number of source samples and \(\|\cdot\|_2\) denotes the L2 norm.

\subsection{Global-Focal Domain Adaptation} 

Most existing fault diagnosis methods primarily align global source and target distributions, while some methods consider local aspects but treat all local perspectives as equally important. However, noise of equivalent intensity can obscure fault-discriminative features differently across categories, leading to class-specific transfer degradation; in other words, each subdomain exhibits a distinct adaptation difficulty. Consequently, we propose a global-focal domain adaptation method in which the global component ensures overall distribution alignment while the focal component adaptively identifies and improves poorly aligned subdomains under noisy conditions. The overall optimization objective of the global-focal domain adversarial module involves domain-adversarial training through GRL, which proceeds as follows:
\begin{equation}  
\begin{aligned}  
    & \underset{\theta_{FR}, \theta_{LC}}{\arg\min} \, \underset{\theta_{GD},\theta_{FD}}{\arg\max} \, E(\theta_{FR}, \theta_{LC}, \theta_{GD},\theta_{FD})= \frac{1}{n_s}\sum_{x_i \in D_s} \mathcal{L}_{LC}(\theta_{FR}, \theta_{LC}) \\
    &  - \lambda \frac{1}{n_s + n_t}\sum_{x_i \in D_s \cup D_t} (\mathcal{L}_{GD}(\theta_{FR}, \theta_{GD})+ \mathcal{L}_{FD}(\theta_{FR}, \theta_{FD}))
\end{aligned}  
\label{eq:adversarial}  
\end{equation}
where $\theta_{LC}$, $\theta_{FD}$, and $\theta_{GD}$ represent the parameters of LC, SDCs, and GDC, respectively, while $\mathcal{L}_{LC}$, $\mathcal{L}_{FD}$, and $\mathcal{L}_{GD}$ denote the corresponding losses.

Since target domain data is unlabeled, we utilize LC outputs as pseudo labels to partition $D_s$ and $D_t$ into $C$ class-specific subdomains $D_s^{(c)}$ and $D_t^{(c)}$, where $c \in \{1, 2, \ldots, C\}$ denotes the class label and the corresponding distributions are $p^{(c)}$ and $q^{(c)}$ respectively. This partitioning enables the construction of $C$ SDCs within the focal domain adaptation component, where each SDC aligns source-target distributions for its respective class $c$. Furthermore, we propose a subdomain attention algorithm (SAA) to prioritize hard-to-transfer focal subdomains by assigning higher attention weights. The focal domain adaptation loss $\mathcal{L}_{FD}$ is formulated as follows:

\begin{equation}
\mathcal{L}_{FD} = \sum_{c=1}^C w_c \cdot \mathcal{L}_c,
\end{equation}
where \( w_c \) represents the subdomain attention weight assigned to class \( c \), and \( \mathcal{L}_c \) denotes the loss associated with class \( c \).

Specifically, for target input $Y_{FR,i}^{t}$, LC generates logits $\mathbf{z}_i^t$ converted via softmax to probability vector $\mathbf{p}_i^t\in\Delta^{C-1}$. Pseudo-label $\tilde{y}_i^t$ is assigned as the max probability class: 
\begin{align}
    \mathbf{p}_i^t &= \sigma(\mathbf{z}_i^t), \quad  \sigma(\mathbf{z}_i^t)_k = \frac{e^{z_{i,k}^t}}{\sum_{j=1}^C e^{z_{i,j}^t}}. \\
    \tilde{y}_i^t &= \arg\max_k \, p_{i,k}^t.
\end{align}
To acquire accurate pseudo-labels that match the ground-truth categories, entropy is also utilized as a filtering threshold. The predictive entropy is defined as follows \citep{sohn2020fixmatch}:
\begin{equation}
\mathcal{H}(\mathbf{p}_i^t) = -\sum_{k=1}^C p_{i,k}^t \log(p_{i,k}^t + \epsilon),
\end{equation}  
where \(\mathcal{H}(\mathbf{p}_i^t)\) represents the predictive entropy, \(p_{i,k}^t\) is the \(k\)-th component of \(\mathbf{p}_i^t\). 

$\tilde{y}_i^t$ is retained only when both confidence (threshold $\xi$) and entropy criteria are satisfied:
\begin{equation}
\max_k p_{i,k}^t > \xi \quad \text{and} \quad \mathcal{H}(\mathbf{p}_i^t) < H_{\text{threshold}},
\end{equation}

We couple the confidence and entropy criteria via an entropy upper bound \citep{sohn2020fixmatch}. For any probability vector with maximum component $m$, define
\begin{equation}
\mathcal{H}_{\max}(m,C) = -\,m\log m - (1-m)\log\frac{1-m}{C-1}.
\end{equation}
Then
\begin{equation}
\mathcal{H}(\mathbf{p}_i^t) \le \mathcal{H}_{\max}(\max_k p_{i,k}^t, C),
\end{equation}
with equality when the non–top-1 mass is uniformly distributed over the remaining $C-1$ classes. To avoid redundancy and couple the thresholds, we set
\begin{equation}
H_{\text{threshold}} = \kappa\, \mathcal{H}_{\max}(\xi,C), \qquad \kappa\in(0,1),
\end{equation}
When $H_{\text{threshold}} \geq \mathcal{H}_{\max}(\xi,C)$, the entropy criterion becomes redundant; otherwise, they are complementary.

Based on these criteria, target and source domain samples sharing the same class are sent to their respective SDCs. The loss \( \mathcal{L}_c \) for class \(c \) is computed using binary cross-entropy:
\begin{equation}  
\mathcal{L}_c = -\frac{1}{n_s^{(c)} + n_t^{(c)}} \left( \sum_{i=1}^{n_s^{(c)}} \log(1 - \hat{d}_i^s) + \sum_{j=1}^{n_t^{(c)}} \log(\hat{d}_j^t) \right). 
\end{equation}  
where \( n_s^{(c)} \) and \( n_t^{(c)} \) are the sample counts for class \( c \) in the source and target domains. \(\hat{d}_i^s\) and \(\hat{d}_j^t\) are the predicted probabilities of source and target samples, respectively.

The detailed implementation of the SAA is presented in Algorithm \ref{alg:dynamic_class_weight}. Noise interference reduces the number of high-quality pseudo-labels, and within a batch, certain subdomains may lack pseudo-labels entirely, leading to training instability and weight allocation failure. To address this issue, the SAA employs an Exponential Moving Average (EMA) \citep{polyak1992acceleration} method to proportionally blend current and historical observations, thereby smoothing cross-batch fluctuations and enhancing trend detection. Additionally, softmax-based weight disparity amplification and sample-size-dependent weight scaling are applied to assign appropriate attention to subdomains that are difficult to align.

GDC functions similarly to DANN \cite{ganin2016domain} in achieving global domain adaptation, using binary cross-entropy as the loss function:
\begin{equation}  
\mathcal{L}_{GD} = -\frac{1}{n_s + n_t} \left( \sum_{i=1}^{n_s} \log(1 - \hat{d}_i^s) + \sum_{j=1}^{n_t} \log(\hat{d}_j^t) \right),  
\end{equation}   
where \( n_s \) and \( n_t \) are the sample counts in the source and target domains, respectively.

\begin{algorithm}[H] 
\caption{Subdomain attention mechanism}  
\label{alg:dynamic_class_weight}  
\begin{algorithmic}[1]  
\REQUIRE Global parameters:
\begin{itemize}[leftmargin=1em, noitemsep, topsep=0pt]
    \item $\alpha \in (0,1)$: Smoothing coefficient;
    \item $\tau > 0$: Temperature parameter;
    \item $m \in (0,1)$: EMA momentum;
    \item $\beta$: Sample count sensitivity coefficient;
    \item $\mathcal{L}_{\text{EMA}} \gets \mathbf{1}_C$ \COMMENT{Cross-batch EMA loss};
    \item $\theta \gets \ln 2$ \COMMENT{Binary CE loss of random guess}.
\end{itemize}

    \FOR{batch $t=1$ to $T$}
        \STATE \textbf{Batch-wise Input ($t$-th batch):}
        \begin{itemize}[leftmargin=0.75cm, noitemsep,topsep=2pt]
            \item \hspace{-1.3em} $C^{(t)}$: Total number of classes;
            \item\hspace{-1.3em} $\mathcal{L}_{\text{c}}^{(t)} \in \mathbb{R}^{C}$: Per-class loss in current batch;
            \item\hspace{-1.3em} $\mathbf{N}_{\text{c}}^{(t)} \in \mathbb{N}^C$: Per-class sample counts in current batch.
        \end{itemize}

    \STATE \textbf{EMA Loss Smoothing:}  
        \STATE $\quad \mathcal{V} \gets \{c \mid N_c > 0\}$ \COMMENT{Classes with samples}; 
        \STATE $\quad \mathcal{U} \gets \{c \mid N_c = 0\}$ \COMMENT{Classes without samples}  ;
        \STATE $\quad \bar{\mathcal{L}} \gets \begin{cases}   
            (1-m)\cdot\mathcal{L}_{\text{c}} + m\cdot\mathcal{L}_{\text{EMA}} & \text{if } N_c > 0 \\
            \mathcal{L}_{\text{EMA}} & \text{if } N_c = 0;
        \end{cases}$  
    \vspace{0.5em}  
        
    \STATE \textbf{Assess Category Alignment Difficulty:}  
        \STATE $\quad \mathbf{d} \gets \max(\theta - \bar{\mathcal{L}}, 0)$  ;
        \STATE $\quad \mathcal{H} \gets \{c \mid d_c > 0\}$ \COMMENT{Under-aligned classes};  
        \STATE $\quad \mathcal{A} \gets \{c \mid d_c = 0\}$ \COMMENT{Well-aligned classes};
    \vspace{0.5em}  
  
    \STATE \textbf{Base Weight Calculation:}  
        \STATE $\quad \mathbf{s} \gets \text{softmax}(\mathbf{d}[\mathcal{H}] / \tau)$;  
        \STATE $\quad u \gets 1/C$ \COMMENT{Uniform prior}  ;
        \STATE $\quad \mathbf{w}[\mathcal{H}] \gets (1-\alpha)\cdot\mathbf{s} + \alpha\cdot u$ ; 
        \STATE $\quad \mathbf{w}[\mathcal{A}] \gets \alpha\cdot u$;
    \vspace{0.5em}
    
    \STATE \textbf{Sample-Size Weight Scaling:}  
        \STATE $\quad \mathbf{q} \gets (\mathbf{N} + \epsilon)^\beta$; 
        \STATE $\quad \mathbf{w}[\mathcal{V}] \gets \mathbf{w}[\mathcal{V}] \circ \mathbf{q}[\mathcal{V}]$;
    \vspace{0.5em}
    
    \STATE \textbf{Weight Normalization:}  
        \STATE  $\quad \mathbf{w_c} \gets \frac{\displaystyle \mathbf{w}}  { \sum \displaystyle\mathbf{w} + \epsilon}$;
    \vspace{0.5em}

    \STATE \textbf{Update EMA (batch-wise):}  
        \STATE $\quad \mathcal{L}_{\text{EMA}}[\mathcal{V}] \gets m \cdot\mathcal{L}_{\text{EMA}}[\mathcal{V}] + (1-m)\cdot\mathcal{L}_{\text{c}}[\mathcal{V}]$;
        \STATE $\quad \mathcal{L}_{\text{EMA}}[\mathcal{U}] \gets \mathcal{L}_{\text{EMA}}[\mathcal{U}]$ \COMMENT{Frozen};
    \vspace{0.5em}

    \STATE \textbf{Output Batch Weights:} $\mathbf{w_c}^{(t)} \in \mathbb{R}^C$. 
    \ENDFOR
\end{algorithmic}  
\end{algorithm}

Global-focal domain adaptation theoretically models the alignment of both marginal and conditional distributions. According to Ben-David \citep{ben2006analysis}, the target domain error is bounded by:
\begin{equation} 
\epsilon_t(h) \leq \epsilon_s(h) + d_{\mathcal{H}}(D_s,D_t) + C_0,
\end{equation}   
where the $\mathcal{H}$-divergence $d_{\mathcal{H}}(D_s,D_t)$ decomposes into global divergence between marginal distributions $P_s(x)$ and $P_t(x)$ (aligned by GDC) and local divergence between conditional distributions $P_s(x|y)$ and $P_t(x|y)$ (aligned by SDCs). This provides solid theoretical grounding for our Global-focal architecture.

\begin{figure*}[!htbp]
    \centering
    \includegraphics[width=0.95\linewidth]{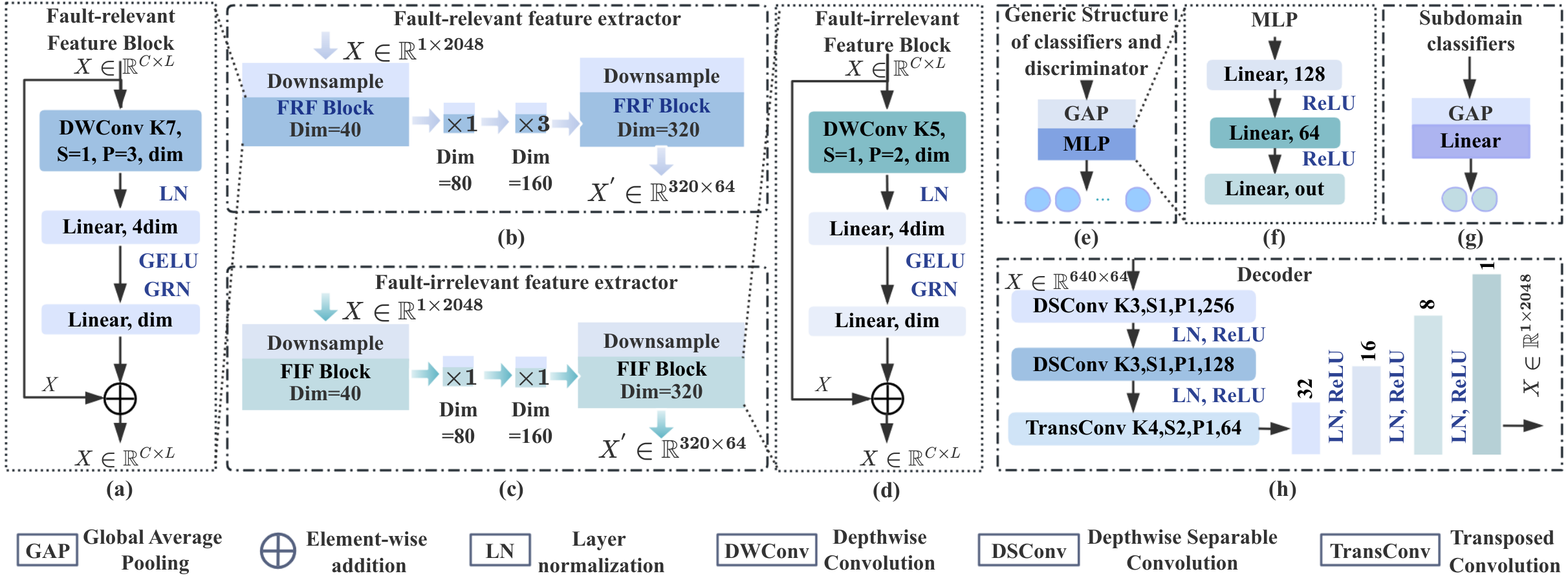}
    \caption{Concrete structure of model components: (a) Fault-relevant feature extraction block; (b) Fault-relevant feature extractor; (c) Fault-irrelevant feature extractor; (d) Fault-irrelevant feature extraction block; (e) Generic structure of classifiers and discriminators; (f) Multi-layer perceptron; (g) Subdomain classifiers; (h) Decoder. }\label{Fig-Concrete-structure}
\end{figure*}

\subsection{Concrete Structure of Model Components} 

The structural design of each module in ISGFAN is shown in Figure \ref{Fig-Concrete-structure}. Both FRFE and FIFE, which have similar architectures, employ four extraction stages, each comprising a downsampling layer and depthwise separable convolution blocks. Details of the downsampling layers and the corresponding changes in tensor dimensions are provided in Table \ref{tab: stages}. For FRFE, the FRF block uses depthwise convolution with a kernel size of $1\times7$ to capture extensive spatial features, followed by layer normalization. Linear layers replace pointwise convolution for mapping cross-channel correlations, enhancing efficiency while preserving mathematical equivalence. These layers expand and compress feature channels, utilizing GELU activation and global response
normalization (GRN) to improve feature extraction quality \citep{woo2023convnext}. Residual connections at the output aid gradient flow and keep the original features \citep{he2016deep}. The FIF block in FIFE implements a similar structure. 

\begin{table}[ht]  
    \centering  
    \small
    \caption{Phases of feature extraction and corresponding dimensions}  
    \begin{tabular}{cccc}  
        \toprule  
        \textbf{Stage} & \textbf{Feature maps} & \textbf{Down-sampling} & \textbf{Channel dims} \\
        \midrule  
        1 & \rule{0pt}{3ex} \( 1 \times \frac{L}{4} \)  & Kernel $1\times 4$, S=4 & 40 \\
        2 & \rule{0pt}{3ex} \( 1 \times \frac{L}{8} \)  & Kernel $1\times 2$, S=2 & 80 \\
        3 & \rule{0pt}{3ex} \( 1 \times \frac{L}{16} \)  & Kernel $1\times 2$, S=2 & 160 \\
        4 & \rule{0pt}{3ex} \( 1 \times \frac{L}{32} \)  & Kernel $1\times 2$, S=2 & 320 \\
        \bottomrule  
    \end{tabular}  
    \label{tab: stages}  
\end{table}
LC, GDC, and LD share the same architecture, with the only difference being in their output dimensions, while SDCs use a single linear layer. The decoder architecture first employs two layers of depthwise separable convolutions with layer normalization to compress and normalize input features, followed by five layers of transposed convolutions for progressive upsampling. 

\subsection{Model Training and Dynamic Loss Weighting} 

In transfer learning-based fault diagnosis methods, a common approach to handling multi-task problems is to assign relatively low base weights to certain losses. However, the effectiveness of this strategy is limited, particularly when numerous losses must be optimized, as gradient domination issues may still occur in multi-task learning. ISGFAN addresses loss imbalance through dynamic weighting during training. In each iteration, auxiliary-to-main loss ratios are monitored, and weights for losses that dominate the gradient flow are adaptively reduced, as detailed in Algorithm \ref{alg: dynamic_Loss_weight}. 

\begin{algorithm}[ht]  
\caption{Dynamic loss weighting}  
\label{alg: dynamic_Loss_weight}  
\begin{algorithmic}[1]  
\REQUIRE ~
    \begin{itemize}[leftmargin=1em, noitemsep, topsep=0pt] 
        \item $\mathcal{L}$: A auxiliary loss for non-classification objectives;  
        \item $\mathcal{L}_{\text{ref}}$: Reference loss (primary classification objective); 
        \item $\lambda_{\text{base}}$: Base weight coefficient;  
        \item $\rho$: The loss threshold ratio for weight reduction. 
    \end{itemize}
    \IF{$\mathcal{L} > \rho \times \mathcal{L}_{\text{ref}}$}  
        \STATE $\lambda \leftarrow \lambda_{\text{base}} \times \dfrac{\rho \times \mathcal{L}_{\text{ref}}}{\,\mathcal{L} + 10^{-18}\,}$;
    \ELSE  
        \STATE $\lambda \leftarrow \lambda_{\text{base}}$;
    \ENDIF  
    \STATE \textbf{return} $\lambda$ 
\end{algorithmic}  
\end{algorithm}

During training, the ISGFAN algorithm optimizes multiple objective loss functions, with trainable parameters for each module updated distinctly based on specific losses, as follows:
\begin{equation}
\begin{aligned}
\theta_{FI} &\leftarrow \theta_{FI}
    - \eta_{FI} \left( \frac{\partial \mathcal{L}_{orth}}{\partial \theta_{FI}}
    + \frac{\partial \mathcal{L}_{recon}}{\partial \theta_{FI}}
    - \frac{\partial \mathcal{L}_{LD}}{\partial \theta_{FI}} \right),\\
\theta_{GD} &\leftarrow \theta_{GD}
    - \eta_{GD} \left( \frac{\partial \mathcal{L}_{GD}}{\partial \theta_{GD}} \right),\\
\theta_{FD} &\leftarrow \theta_{FD}
    - \eta_{FD} \left( \frac{\partial \mathcal{L}_{FD}}{\partial \theta_{FD}} \right),\\
\theta_{LC} &\leftarrow \theta_{LC}
    - \eta_{LC} \left( \frac{\partial \mathcal{L}_{LC}}{\partial \theta_{LC}} \right),\\
\theta_{DE} &\leftarrow \theta_{DE}
    - \eta_{DE} \left( \frac{\partial \mathcal{L}_{recon}}{\partial \theta_{DE}} \right),\\
\theta_{LD} &\leftarrow \theta_{LD}
    - \eta_{LD} \left( \frac{\partial \mathcal{L}_{LD}}{\partial \theta_{LD}} \right),\\
\theta_{FR} &\leftarrow \theta_{FR}
    - \eta_{FR} \Big( \frac{\partial \mathcal{L}_{LC}}{\partial \theta_{FR}}
    + \frac{\partial \mathcal{L}_{orth}}{\partial \theta_{FR}}+ \frac{\partial \mathcal{L}_{recon}}{\partial \theta_{FR}} \\
&\qquad\qquad
    - \frac{\partial \mathcal{L}_{GD}}{\partial \theta_{FR}}
    - \frac{\partial \mathcal{L}_{FD}}{\partial \theta_{FR}} \Big).
\end{aligned}
\end{equation}
where $\theta_{*}$ and $\eta_{*}$ denote the parameters and learning rate for each module, respectively. The total loss during model training is calculated as follows:
\begin{equation}  
\mathcal{L}=\mathcal{L}_{LC}+\delta\mathcal{L}_{GD}+\zeta\mathcal{L}_{FD}+\gamma\mathcal{L}_{orth}+\mu\mathcal{L}_{recon}+\omega\mathcal{L}_{LD},
\label{eq: total_loss}  
\end{equation} 
where \(\delta, \zeta, \gamma, \mu, \omega\) are dynamic weight parameters for each corresponding loss.

\section{Experiment and Results Analysis}
\label{sec: Experiment and Analysis}
This section details the experimental design and presents case studies using three public benchmark datasets, demonstrating the robustness of the proposed ISGFAN method in transfer fault diagnosis under noise interference.

\begin{figure*}[!htbp] 
    \centering
   \includegraphics[width=1\linewidth]{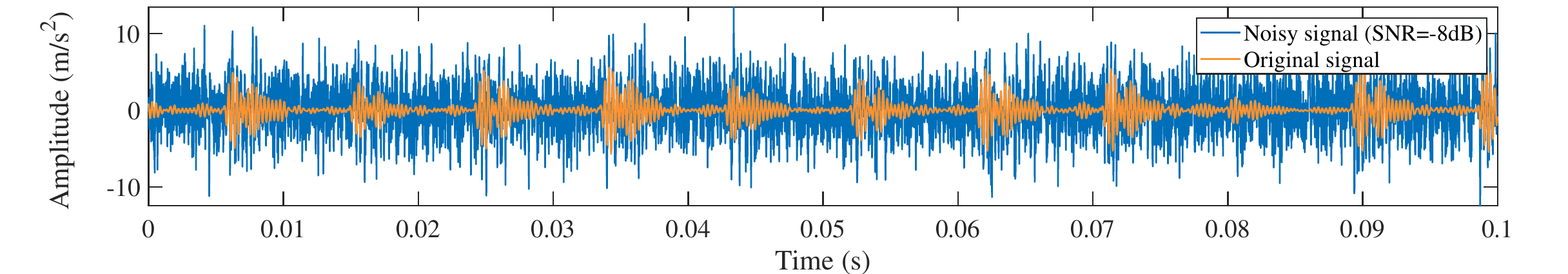}
    \caption{Comparison of raw vibration signal and noise-contaminated vibration signal in the time domain.}
    \label{fig: Samples_signals}
\end{figure*}

\subsection{Experimental Setup}
\subsubsection{Model Training Configuration}

All experiments were conducted using the PyTorch framework based on Python 3.9. The hardware setup included a 14th-generation Intel® Core™ i9 processor and NVIDIA® GeForce RTX 4060. Detailed training configurations of ISGFAN are summarized in Table \ref{tab:training_configuration}. The hyperparameters $\alpha, \tau, m, \beta$ in Algorithm \ref{alg:dynamic_class_weight} were assigned values of 0.05, 0.02, 0.3, and -0.1, respectively. The hyperparameters $\delta, \zeta, \gamma, \mu, \omega, \rho$ in Algorithm \ref{alg: dynamic_Loss_weight} and Eq. \ref{eq: total_loss} were set to 0.5, 0.1, 0.01, 0.01, 0.01, and 10, respectively. These values were empirically determined and validated through experimental testing, representing a standard configuration.

\begin{table}[!t]  
\small 
\caption{Training configuration for the ISGFAN}  
\setlength{\tabcolsep}{4pt}  
\begin{tabular*}{\columnwidth}{@{\extracolsep{\fill}}ll}  
\toprule  
 \textbf{Hyperparameters} & \textbf{Values} \\  
\midrule  
Training Epochs & 3500 \\  
Batch Size & 32 \\  
Base Learning Rate & $1.0 \times 10^{-4}$ \\ 
Minimum Learning Rate & $1.0 \times 10^{-6}$ \\
Learning Rate Scheduler & CosineAnnealingLR \\  
Optimizer & AdamW \\   
Weight Decay & $5.0 \times 10^{-4}$ \\  
\bottomrule  
\end{tabular*}  
\label{tab:training_configuration}  
\end{table}  

\subsubsection{Data Preprocessing}

This study evaluates model robustness using Gaussian noise, Laplacian noise, and mixed noise. Gaussian noise represents normally distributed random variations, Laplacian noise models impulse-like disturbances through its heavy-tailed distribution, and mixed noise, a hybrid of both, exhibits enhanced disturbance capabilities. Let $s_{raw}(t)$ denote the original vibration signal, with the sensor-captured noisy signal $s_{noisy}(t)$ expressed as:  
\begin{equation}  
s_{noisy}(t) = s_{raw}(t) + noise(t)  
\label{eq:noise_addition}  
\end{equation}  
where $noise(t)$ represents the three noise types. Noise intensity is controlled by the signal-to-noise ratio (SNR) \citep{zhou2023rotating}:  
\begin{equation}  
SNR_{dB} = 10 \log_{10} \left( \frac{P_{signal}}{P_{noise}} \right)  
\end{equation}  
where $P_{signal}$ and $P_{noise}$ denote the power of the original signal and noise, respectively. A noise level of SNR = -8 dB is adopted for severe interference. Figure \ref{fig: Samples_signals} compares raw and noise-contaminated vibration signals; at SNR = -8 dB, noise overwhelms fault pulses, making characteristic features barely discernible.

After adding noise, the source and target domain data are organized according to operating condition categories. The training set comprises all source domain data and unlabeled target domain data, while the test set comprises the labeled target domain data.

\subsubsection{Comparison Method}
As indicated in the literature, noise-resistant fault diagnosis approaches and transfer learning-based methods are typically studied separately. For this study, six advanced methods were selected for comparison:

\textbf{Noise-resistant fault diagnosis methods:}
\begin{enumerate}[topsep=3pt,itemsep=1pt,parsep=0pt]
    \item Multi-timescale Attention Residual Shrinkage Network (AMARSN) \cite{gao2024multi1}: This method employs multi-scale convolution and attention mechanisms for adaptive denoising.
    \item Noise-Resilient Optimized Residual Network (NORN) \cite{chen2025noise}: This approach proposes inverted residual structures and soft-thresholding functions for filtering and denoising.
    \item Adaptive Graph Framelet Convolutional Network (AGFCN) \cite{he2025agfcn}: This network designs graph convolution based on framelet transform for denoising.
\end{enumerate}

\textbf{Transfer fault diagnosis methods:}
\begin{enumerate}[topsep=3pt,itemsep=1pt,parsep=0pt,resume]
    \item Deep Subdomain Adaptation Network (DSAN) \cite{zhu2020deep}: This method utilizes entropy and Maximum Mean Discrepancy (MMD) for subdomain-specific alignment.
    \item Parallel Ensemble Optimization Transfer Fault Diagnosis Framework (PEOTL) \cite{tang2024parallel}: This framework features a transfer learning network with low-dimensional feature reuse and develops a parallel ensemble optimization loss function.
    \item Indirect Transfer Fault Diagnosis method (ITFS) \cite{qian2024new}: This method employs adversarial approaches and introduces Gaussian priors for indirect closed-set domain adaptation.
\end{enumerate}

Accuracy was employed as the evaluation metric, with all experiments repeated five times to ensure statistical reliability. The reported accuracy values represent the mean of the obtained results. Accuracy \citep{powers2011evaluation} is calculated as follows:
\begin{equation}
Accuracy = \frac{TP + TN}{TP + TN + FP + FN}
\label{eq:24}
\end{equation}
Where $TP$, $TN$, $FP$, and $FN$ denote the counts of true positives, true negatives, false positives, and false negatives, respectively. Data and code are available at \url{https://github.com/JYREN-Source/ISGFAN}

\subsection{Case Study I }
\subsubsection{Dataset Description}
\begin{figure} 
    \centering
    \includegraphics[width=0.9\linewidth]{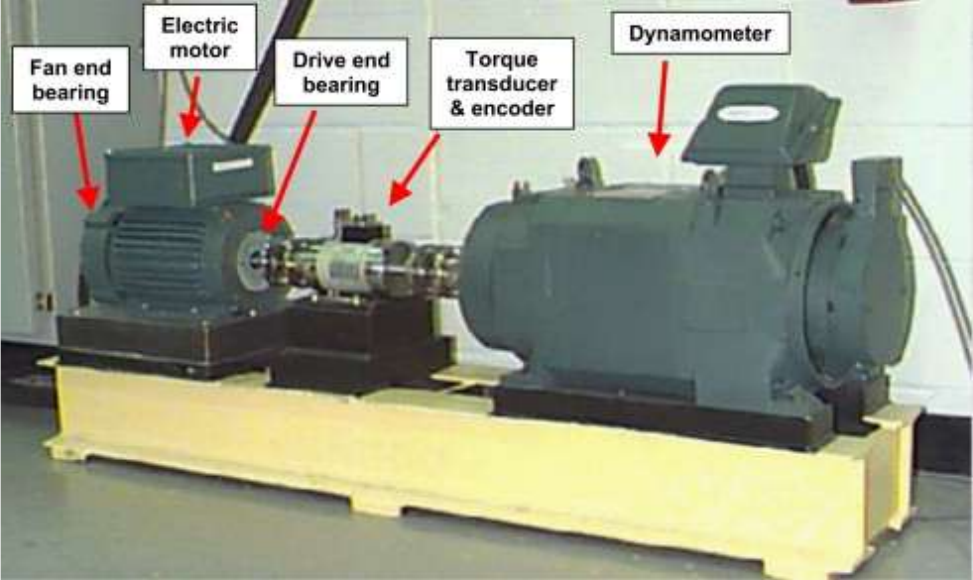}
    \caption{Test bench setup for CWRU \cite{smith2015rolling}: An overview of the system used for vibration testing and data acquisition.}
    \label{fig:CWRU}
\end{figure}

\begin{table}[!t]  
\small
\caption{Fault types and sample counts of vibration data under 1, 2, and 3 HP load conditions in the CWRU dataset.}  
\setlength{\tabcolsep}{0pt}    
\begin{tabular*}{\columnwidth}{@{\extracolsep{\fill}}lccc}  
\toprule  
\textbf{Fault type} & \textbf{Damage size (inch)} & \textbf{Label} & \textbf{Total samples} \\
\midrule  
Inner fault & 0.007 & I1 & 210 \\
            & 0.014 & I2 & 210 \\
            & 0.021 & I3 & 210 \\[0.5ex]  
Outer fault & 0.007 & O1 & 210 \\
            & 0.014 & O2 & 210 \\
            & 0.021 & O3 & 210 \\[0.5ex]  
Ball fault  & 0.007 & B1 & 210 \\
            & 0.014 & B2 & 210 \\
            & 0.021 & B3 & 210 \\[0.5ex]  
Normal      & --    & N  & 210 \\
\bottomrule  
\end{tabular*}  
\label{tab:load_condition_CWRU}  
\end{table}
The CWRU rolling bearing dataset, a renowned benchmark in rotating machinery fault diagnosis \cite{smith2015rolling}, is provided by the Case Western Reserve University and is available at \url{https://engineering.case.edu/bearingdatacenter}. The dataset comprises vibration signals collected from the test platform illustrated in Figure \ref{fig:CWRU}. The tested bearing, model SKF 6205-2RS JEM, simulates four distinct health conditions: normal, inner ring fault, ball fault, and outer ring fault. Various bearing faults, characterized by damage sizes of 0.007, 0.014, and 0.021 inches, were introduced via electric discharge machining. The dataset creators applied different loads (0, 1, 2, and 3 HP) to the bearings, with a sampling frequency of 48 kHz. Under these operating conditions, vibration signals were collected from the drive end of the bearing. 

For this experiment, vibration data from the CWRU dataset were utilized under 1, 2, and 3 HP load conditions. In subsequent experimental results, the transfer task '1-2' indicates training on data from the 1 HP load condition and testing on data from the 2 HP load condition; a similar approach applies to the other tasks. During data preprocessing, a segment length of 2048 points was defined as a sample for model training or testing. Table \ref{tab:load_condition_CWRU} presents the specific categories and total sample counts of the vibration data collected under these three load conditions.

\subsubsection{Test Results and Analysis}

\begin{table}[!t]  
    \small
    \caption{Comparison of model sizes and training durations for the analyzed models.}  
\setlength{\tabcolsep}{0pt}  
\begin{tabular*}{\columnwidth}{@{\extracolsep{\fill}}ccc}  
\toprule  
\textbf{Model} & \textbf{Parameters} & \textbf{Training time (s)} \\
\midrule  
AMARSN           & 0.03 M & 627.5 \\
NORN         & 0.42 M & 235.6 \\
AGFCN  & 1.26 M & 705.2 \\
DSAN  & 4.57 M & 3121.8 \\
ITFS     & 4.64 M & 6821.4 \\
PEOTL      & 7.13 M & 4723.2 \\
ISGFAN  & 3.41 M & 4601.4 \\
\bottomrule  
\end{tabular*}  
\label{tab:model_comparison}  
\end{table}

\begin{table*}[!htbp]
    \small
    \caption{Transfer accuracy of ISGFAN on the CWRU dataset across various mixed noise levels.}
    \setlength{\tabcolsep}{4pt} 
\begin{tabular*}{\textwidth}{@{\extracolsep{\fill}} c l | *{7}{c} }
\toprule
\textbf{Noise type} & \multicolumn{1}{c}{\textbf{Intense level}} & \textbf{1-2} & \textbf{1-3} & \textbf{2-1} & \textbf{2-3} & \textbf{3-1} & \textbf{3-2} & \textbf{Average} \\
\midrule
\multirow{6}{*}{Mixed}
& 0 dB  & $97.51\%$  & $99.15\%$ & $98.24\%$  & $97.89\%$  & $99.02\%$  & $97.86\%$  & $98.28\%$ \\
& -2 dB & $96.26\%$  & $98.83\%$ & $96.02\%$  & $97.32\%$  & $98.65\%$  & $97.06\%$  & $97.36\%$ \\
& -4 dB & $93.11\%$  & $98.52\%$ & $94.12\%$  & $95.22\%$  & $97.35\%$  & $94.06\%$  & $95.40\%$ \\
& -6 dB & $90.23\%$  & $96.85\%$ & $91.72\%$  & $91.31\%$  & $96.43\%$  & $89.06\%$  & $92.60\%$ \\
& -8 dB & $81.81\%$  & $95.48\%$ & $88.40\%$  & $88.24\%$  & $94.07\%$  & $83.16\%$  & $88.53\%$  \\ 
\bottomrule
\end{tabular*}
\label{tab:diagnostic_accuracy_CWRU}
\end{table*}

\begin{table*}[t]
\small
\footnotesize
\caption{Comparison of models' performance: Transfer accuracy of each method on the CWRU dataset across various tasks.}
\setlength{\tabcolsep}{3pt}  
\renewcommand{\arraystretch}{0.75}
\begin{tabular*}{\textwidth}{@{\extracolsep{\fill}} l >{\arraybackslash}p{2cm} | *{7}{c} }  
\toprule
\textbf{Noise type} & \multicolumn{1}{l}{\textbf{Model}} & \textbf{1-2} & \textbf{1-3} & \textbf{2-1} & \textbf{2-3} & \textbf{3-1} & \textbf{3-2} & \textbf{Average} \\
\midrule
\multirow{7}{*}{Gaussian (-8 dB)}
& AMARSN  & $72.89\%$ & $82.42\%$ & $80.83\%$ & $82.08\%$ & $84.79\%$ & $69.94\%$ & $78.66\%$ \\
& NORN    & $71.78\%$ & $77.93\%$ & $80.81\%$ & $78.12\%$ & $79.95\%$ & $72.47\%$ & $76.84\%$ \\
& AGFCN   & $68.83\%$ & $77.29\%$ & $74.41\%$ & $75.28\%$ & $77.33\%$ & $67.95\%$ & $73.52\%$ \\
& ITFS    & $75.56\%$ & $85.97\%$ & $85.49\%$ & $86.84\%$ & $87.74\%$ & $77.71\%$ & $83.22\%$ \\
& DSAN    & $80.35\%$ & $91.28\%$ & $87.42\%$ & $87.63\%$ & $91.84\%$ & $81.36\%$ & $86.65\%$  \\
& PEOTL   & $73.72\%$ & $84.19\%$ & $82.91\%$ & $84.96\%$ & $86.98\%$ & $70.96\%$ & $80.62\%$ \\
& \textbf{ISGFAN}  & $\textbf{82.95\%}$ & $\textbf{97.58\%}$ & $\textbf{89.47\%}$ & $\textbf{90.23\%}$ & $\textbf{96.00\%}$ & $\textbf{84.43\%}$ & $\textbf{90.11\%}$ \\
\cmidrule(l){2-9}  
\multirow{7}{*}{Laplacian (-8 dB)}
& AMARSN  & $71.70\%$ & $80.94\%$ & $79.46\%$ & $80.78\%$ & $83.71\%$ & $68.86\%$ & $77.58\%$ \\
& NORN    & $70.36\%$ & $76.82\%$ & $79.72\%$ & $77.03\%$ & $78.80\%$ & $71.21\%$ & $75.66\%$ \\
& AGFCN   & $67.60\%$ & $75.90\%$ & $73.31\%$ & $74.02\%$ & $76.03\%$ & $68.93\%$ & $72.63\%$ \\
& ITFS    & $74.41\%$ & $84.92\%$ & $84.15\%$ & $85.62\%$ & $86.68\%$ & $76.46\%$ & $82.04\%$ \\
& DSAN    & $77.54\%$ & $90.14\%$ & $86.03\%$  &$86.19\%$ & $90.32\%$  &$80.09\%$ & $85.05\%$ \\
& PEOTL   & $72.45\%$ & $83.10\%$ & $81.43\%$ & $83.57\%$ & $85.51\%$ & $69.51\%$ & $79.26\%$ \\
& \textbf{ISGFAN}  & $\textbf{81.76\%}$ & $\textbf{96.14\%}$ & $\textbf{88.06\%}$ & $\textbf{89.05\%}$ & $\textbf{94.70\%}$ & $\textbf{83.09\%}$ & $\textbf{88.80\%}$ \\
\cmidrule(l){2-9}  
\multirow{7}{*}{Mixed (-8 dB)}
& AMARSN  & $70.83\%$ & $79.55\%$ & $78.23\%$ & $79.37\%$ & $82.77\%$ & $66.97\%$ & $76.29\%$ \\
& NORN    & $69.35\%$ & $75.64\%$ & $78.20\%$ & $75.98\%$ & $77.66\%$ & $70.10\%$ & $74.49\%$ \\
& AGFCN   & $66.22\%$ & $75.12\%$ & $72.34\%$ & $72.33\%$ & $74.36\%$ & $65.14\%$ & $70.92\%$ \\
& ITFS    & $73.53\%$ & $83.06\%$ & $83.23\%$ & $84.18\%$ & $85.43\%$ & $75.19\%$ & $80.77\%$ \\
& DSAN    & $74.81\%$ & $89.32\%$ & $81.41\%$ & $83.26\%$ & $87.11\%$ & $78.16\%$ & $82.35\%$ \\
& PEOTL   & $71.12\%$ & $81.27\%$ & $80.82\%$ & $82.76\%$ & $84.93\%$ & $68.63\%$ & $78.26\%$ \\
& \textbf{ISGFAN}  & $\textbf{81.81\%}$ & $\textbf{95.48\%}$ & $\textbf{88.40\%}$ & $\textbf{88.24\%}$ & $\textbf{94.07\%}$ & $\textbf{83.16\%}$ & $\textbf{88.53\%}$ \\
\cmidrule(l){2-9}
\bottomrule
\end{tabular*}
\label{tab: comparison_accuracy_CWRU}
\end{table*}

\begin{figure*}[!htbp] 
\centering
\includegraphics[width=0.95\linewidth]{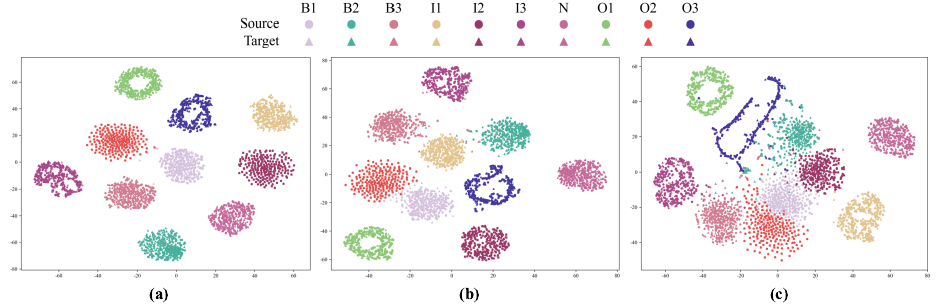}
\caption{t-SNE visualization of ISGFAN on CWRU 1-3 transfer task under mixed noise at varying SNRs: (a) 0 dB; (b) -4 dB; (c) -8 dB.} \label{fig: Fig_tsne1_CWRU}
\end{figure*}

Table \ref{tab:diagnostic_accuracy_CWRU} presents the transfer accuracy of the proposed ISGFAN method on the CWRU dataset under mixed-noise interference. At an SNR of 0 dB, ISGFAN achieves an average accuracy of 98.28\%, while maintaining an accuracy of 88.53\% even when the SNR drops to -8 dB, demonstrating its robust noise-resistant and transferable fault diagnosis capabilities. Notably, for the transfer tasks 3-1 and 3-2, ISGFAN attains accuracies of 99.02\% and 97.86\%, respectively, at SNR = 0 dB. However, at SNR = -8 dB, the accuracies drop to 94.07\% and 83.16\%, respectively, with task 3-2 exhibiting higher noise sensitivity. This highlights the complex interplay between noise and domain shift in the performance of fault diagnosis.  

Figure \ref{fig: Fig_tsne1_CWRU} illustrates the t-SNE visualizations of ISGFAN for the 1-3 transfer task under 0 dB, -4 dB, and -8 dB conditions. The results reveal that ISGFAN achieves excellent distribution alignment when the noise power matches the original signal power (0 dB). As the noise power increases, some categories begin to exhibit confusion. Under the extreme -8 dB mixed-noise condition, ISGFAN maintains relatively distinct discriminative boundaries. These findings highlight the significant impact of noise on classification and transfer performance while demonstrating ISGFAN's robustness for cross-domain applications under signal degradation conditions.  

\begin{figure*}[!htbp] 
\centering
\includegraphics[width=1\linewidth]{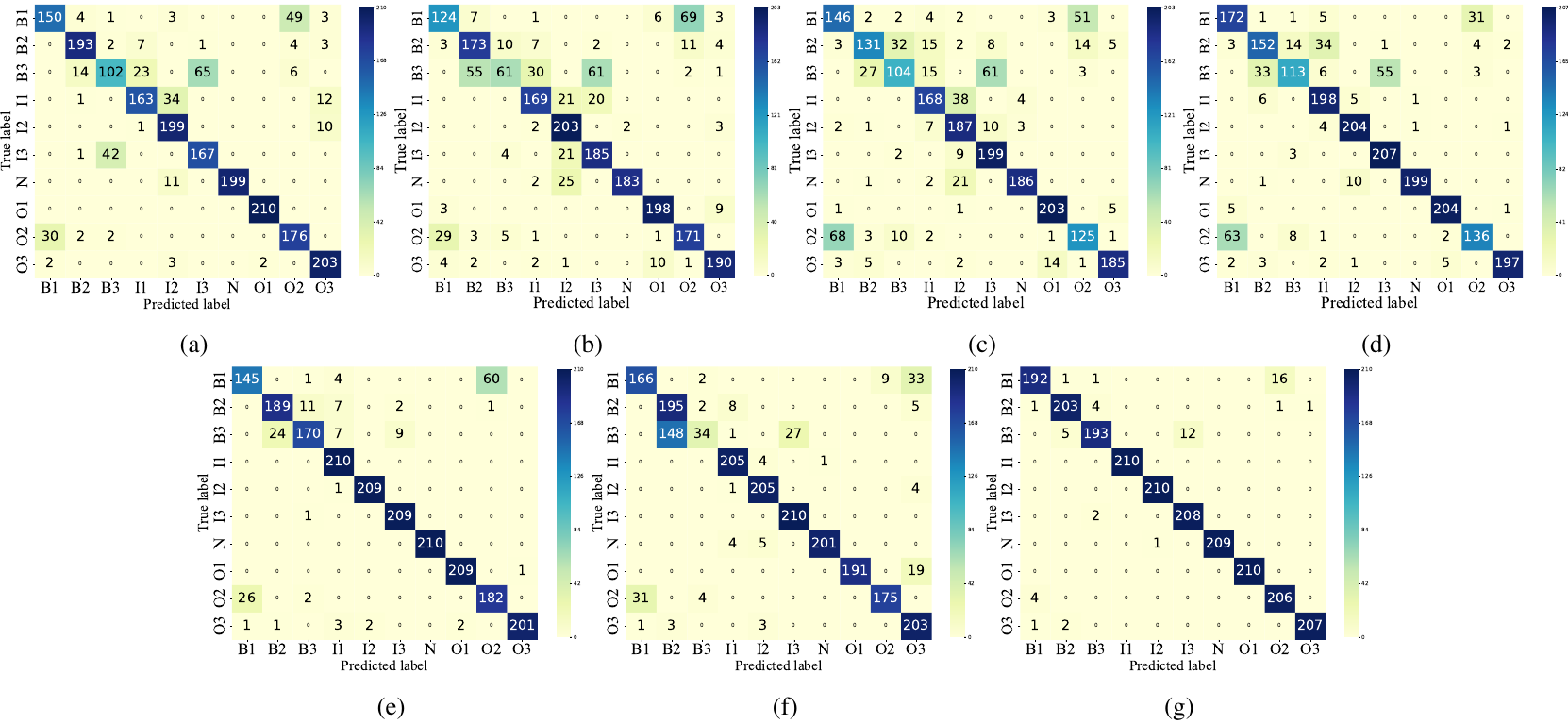}
\caption{Confusion matrices of each method tested on the CWRU 1-3 task under Gaussian noise with an SNR of -8 dB: (a) AMARSN ; (b) NORN; (c) AGFCN; (d) PEOTL; (e) DSAN; (f) ITFS; (g) ISGFAN.} \label{fig: Fig_CM_CWRU}
\end{figure*}
Table \ref{tab: comparison_accuracy_CWRU} demonstrates ISGFAN's superior performance compared to other advanced models on the CWRU dataset. Under -8 dB noise conditions, ISGFAN consistently achieves the highest accuracy with averages of 90.11\%, 88.80\%, and 88.53\% for Gaussian, Laplacian, and Mixed noise, respectively. ISGFAN outperforms the second-best method, DSAN, by 3.45-6.18\% and surpasses ITFS by 6.76-7.76\% across different noise types. Moreover, Table \ref{tab:model_comparison} shows that both DSAN and ITFS require significantly more parameters than ISGFAN, demonstrating the efficiency of the proposed method. In challenging transfer scenarios (1-2 and 3-2), ISGFAN maintains over 81\% accuracy under severe mixed noise, while achieving excellent performance exceeding 94\% in favorable tasks (1-3 and 3-1).  Noise-resilient models focus on noise suppression without incorporating domain adaptation mechanisms, resulting in poor cross-domain performance. Domain adaptation methods outperform noise-resistant models in transfer scenarios but remain vulnerable to severe noise interference. These results highlight the necessity for solutions that simultaneously achieve robust noise resistance and effective cross-domain transfer, which ISGFAN successfully addresses.

Confusion matrices were constructed to quantitatively evaluate the classification performance of each method, as illustrated in Figure \ref{fig: Fig_CM_CWRU}. ISGFAN achieved an average diagnostic accuracy of 90\% across all fault categories, demonstrating outstanding performance in categories I1, I2, I3, N, and O1, with classification accuracies exceeding 99\% for each category. Comparative analysis of subplots (a) through (f) reveals that ISGFAN substantially reduces misclassifications across various fault categories. The primary confusion occurs between fault pairs B1-O2 and B3-I3, a phenomenon observed across all comparative models. This is likely attributed to significant noise interference that disrupts the distinguishing features of these specific fault categories, posing challenges for classification and transfer learning. Furthermore, other comparative models exhibit various category-specific confusions. These results indicate that the proposed method achieves superior performance in the presence of noise interference, providing a reliable technical foundation for industrial applications.

\begin{figure*}[!htbp] 
\centering
\includegraphics[width=1\linewidth]{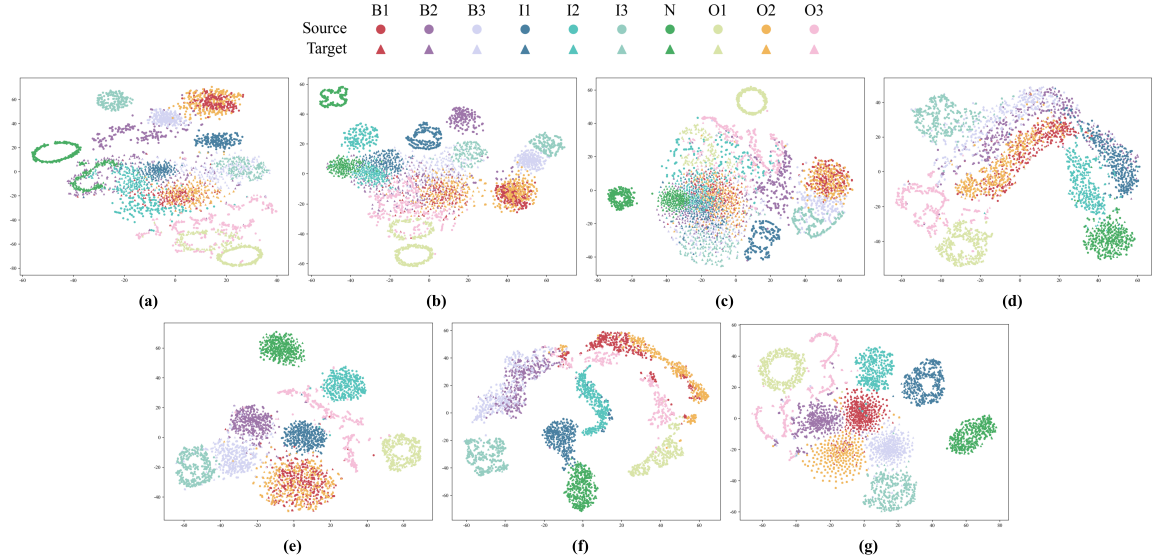}
\caption{t-SNE of each method tested on the CWRU 3-1 task under Laplacian noise with an SNR of -8 dB: (a) AMARSN ; (b) NORN; (c) AGFCN; (d) PEOTL; (e) DSAN; (f) ITFS; (g) ISGFAN. }\label{Fig-tsne}
\end{figure*}

Figure \ref{Fig-tsne} presents t-SNE visualization analysis of each method on task 3-1 under -8 dB Laplacian noise conditions. Subplots (a)-(c) show that the three noise-resistant models generate chaotic scatter plots where, despite clear source domain clustering, target domain data of identical categories significantly deviate from source domain counterparts. This demonstrates that relying solely on noise-resistant fault diagnosis models is insufficient to address cross-domain fault diagnosis problems simultaneously. Subplots (d)-(f) show that PEOTL and ITFS generate point clusters that begin to converge in different directions, with distinct distribution characteristics of various categories. DSAN produces clearer clustering boundaries, but the B1 and O3 categories are almost confused together. All three methods struggle to separate multiple categories in latent space into clearly defined point clusters, highlighting their limitations under strong noise interference and their inability to address the indirect domain gap expansion caused by noise. According to the comparison between subplots (g) and subplots (d)-(f) in Figure \ref{Fig-tsne}, it can be determined that ISGFAN significantly increases the distribution distance of inter-class samples in the latent space, with each category having relatively clear decision boundaries. The information separation architecture effectively minimizes interference from noise and fault-irrelevant information, enabling the model to focus on domain-invariant fault representations. The global-focal domain adversarial approach comprehensively enhances distribution alignment.

\subsubsection{ Ablation Study and Validation of Effectiveness }

Ablation experiments are conducted on 2-3 transfer tasks under -8 dB Gaussian noise conditions to verify the effectiveness of the proposed innovative modules. Table \ref{tab: ablation} presents the ablation study results. ISGFAN-ISFA represents the configuration in which both the information separation architecture and focal domain adversarial module are removed from ISGFAN, making it architecturally equivalent to DANN. ISGFAN-IS corresponds to the ISGFAN variant with the information separation architecture removed, while ISGFAN-FA denotes the configuration with the focal domain adversarial module removed. ISGFAN-FALD represents ISGFAN with both the focal domain adversarial module and the LD removed. Experimental results demonstrate that all components contribute to overall performance enhancement, with the focal domain adversarial module exhibiting the most substantial impact: ISGFAN-IS achieves a 6.68$\%$ accuracy improvement over the baseline model, whereas the implementation of the information separation architecture delivers a significant 4.91$\%$ performance gain. LD also proves effective, as its removal results in a 1.96$\%$ accuracy reduction, validating that adversarial training can help the model more sufficiently separate fault-irrelevant features. The complete ISGFAN architecture achieves optimal performance at 90.23$\%$ accuracy. 

\begin{figure}
\centering
\includegraphics[width=0.9\linewidth]{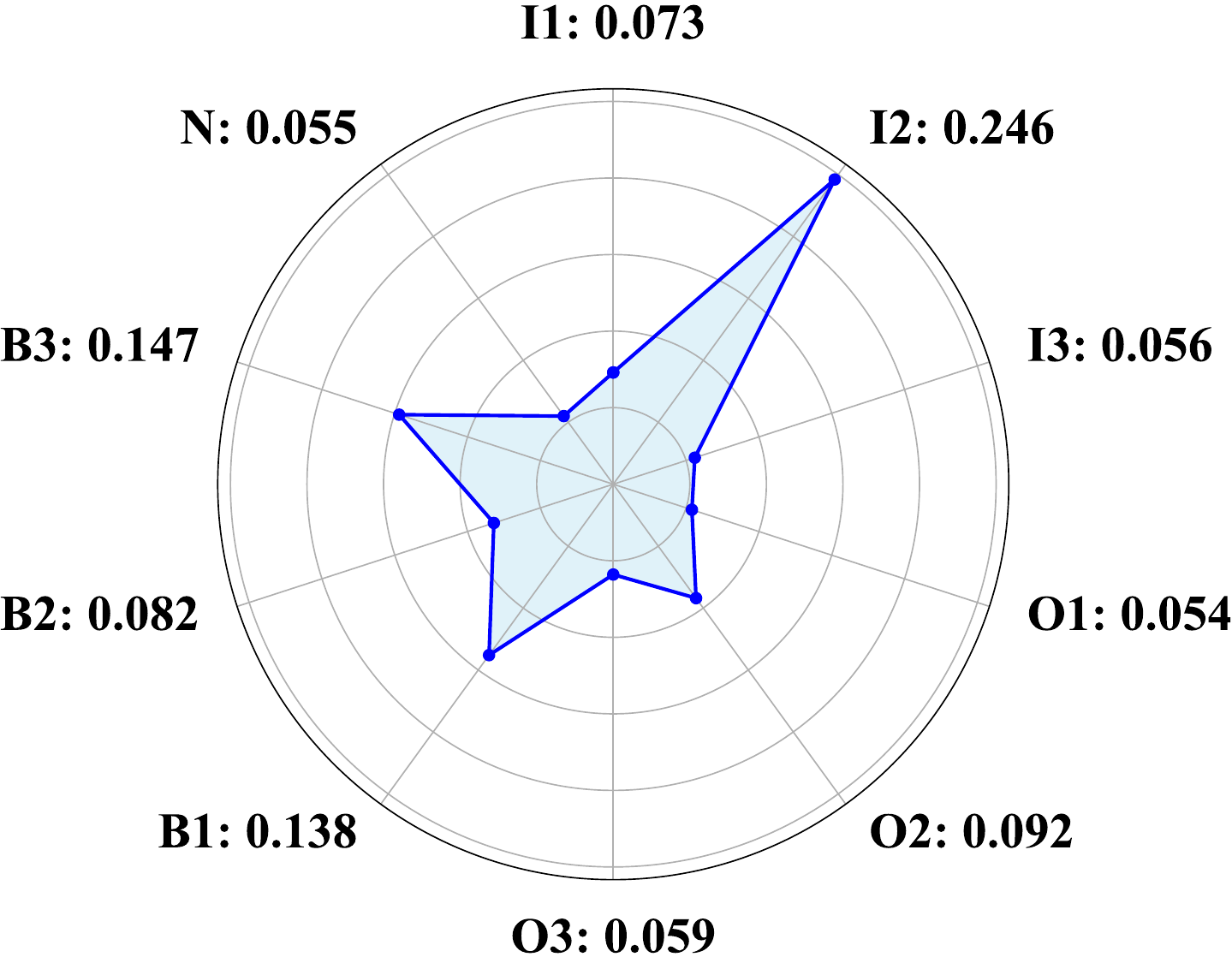}
\caption{Category-wise attention weights assigned by SAA}\label{fig: weights_radar}
\end{figure}

\begin{table}[!htbp]  
\small 
\caption{Result of ablation study}  
\label{tab: ablation}  
\centering
\setlength{\tabcolsep}{15pt}
\begin{tabular}{c c}  
\toprule  
\textbf{Model} & \textbf{Average accuracy}  \\  
\midrule  
\addlinespace[1.5pt]
ISGFAN-ISFA & 81.33\% \\  
ISGFAN-IS & 88.01\% \\  
ISGFAN-FA & 86.24\% \\  
ISGFAN-FALD & 84.28\% \\  
ISGFAN & 90.23\%\\  
\bottomrule  
\end{tabular}  
\end{table}

\begin{figure*}
\centering
\includegraphics[width=1\linewidth]{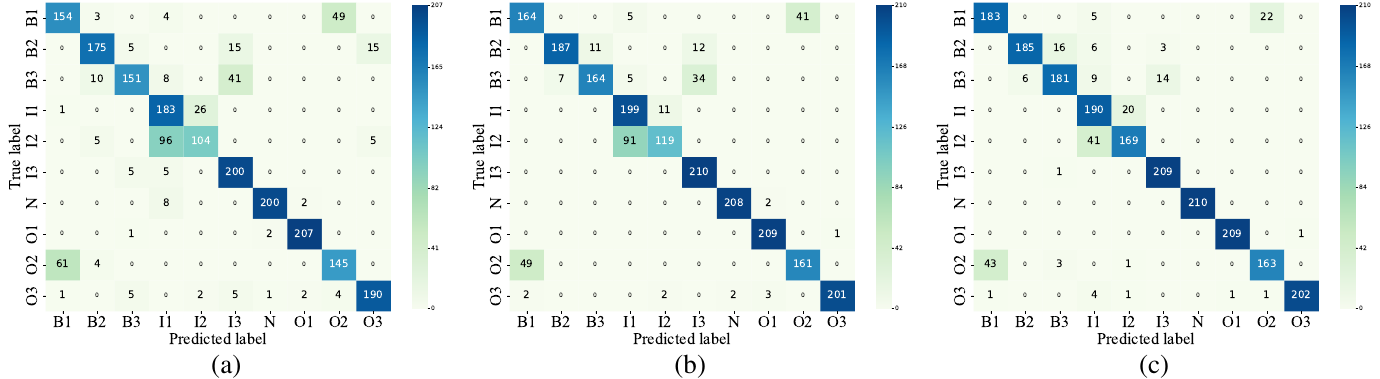}
\caption{The confusion matrices resulting from the ablation study: (a) ISGFAN-ISFA ; (b) ISGFAN-FA; (c) ISGFAN.}\label{fig: Fig_CM2_CWRU}
\end{figure*}
Figure \ref{fig: Fig_CM2_CWRU} demonstrates the efficacy of the proposed architecture through confusion matrices. A comparison between subfigures (a) and (b) reveals that integrating the information separation architecture significantly reduces misclassification rates. Specifically, for categories exhibiting strong transferability (I1, I3, N, O1, and O3), ISGFAN-ISFA still misclassifies certain samples, whereas the information separation architecture effectively resolves these ambiguities. For categories with poor transfer performance (B1, B3, I2, O2), the architecture also yields consistent improvements. These results validate the superiority of the information separation architecture, where the information separation guidance branch effectively assists the main branch in focusing on fault-related domain-invariant features through improved orthogonality loss and adversarial training, thereby isolating the interference of noise and domain-specific information during the transfer process. However, ISGFAN-FA exhibits notable limitations: B1, B3, and I2 still retain a substantial number of misclassified samples. The incorporation of SDC and the SAA addresses these shortcomings. A comparison between subfigures (b) and (c) demonstrates significant accuracy improvements for B1, B3, and I2. Figure \ref{fig: weights_radar} depicts the attention weights assigned by the SAA to each category during training (averaged over all iterations). The mean weights for B1, B3, and I2 are 0.138, 0.148, and 0.246, respectively, corresponding to increases of 19, 17, and 50 correctly classified samples, affirming the efficacy of the focal domain adaptation strategy. The EMA algorithm embedded in SAA alleviates instability in pseudo-label quality and quantity, preventing model misguidance. By synergizing the pseudo-label strategy with the SAA, SDC optimizes local distribution alignment for challenging transfer categories under noisy, unsupervised conditions. The Global-focal Domain Adversarial module, comprising SDC and GDC, effectively constrains both marginal and conditional distributions, achieving robust transfer performance under noise interference. 

\subsection{Case Study II}
\subsubsection{Dataset Description} 

\begin{table}[!htbp]  
    \small
    \caption{Categories of vibration data and sample size across load conditions}  
\setlength{\tabcolsep}{0pt}    
\begin{tabular*}{\columnwidth}{@{\extracolsep{\fill}}lccc}  
\toprule  
\textbf{Fault type} & \textbf{Damage size} & \textbf{Label} & \textbf{Total samples} \\
\midrule   
Inner fault & 0.3 mm & I1 & 600 \\
            & 1.0 mm & I2 & 600 \\
            & 3.0 mm & I3 & 600 \\[0.5ex]  
Outer fault & 0.3 mm & O1 & 600 \\
            & 1.0 mm & O2 & 600 \\
            & 3.0 mm & O3 & 600 \\[0.5ex]  
Misalignment & 0.1 mm & M1 & 600 \\
             & 0.3 mm & M2 & 600 \\
             & 0.5 mm & M3 & 600 \\[0.5ex]  
Unbalance   & 583 mg  & U1 & 600 \\
            & 1751 mg & U2 & 600 \\
            & 3318 mg & U3 & 600 \\[0.5ex]  
Normal      & --      & N  & 600 \\
\bottomrule  
\end{tabular*}  
\label{tab:vibration_data_KAIST} 
\end{table}

\begin{figure} 
        \centering
        \includegraphics[width=1\linewidth]{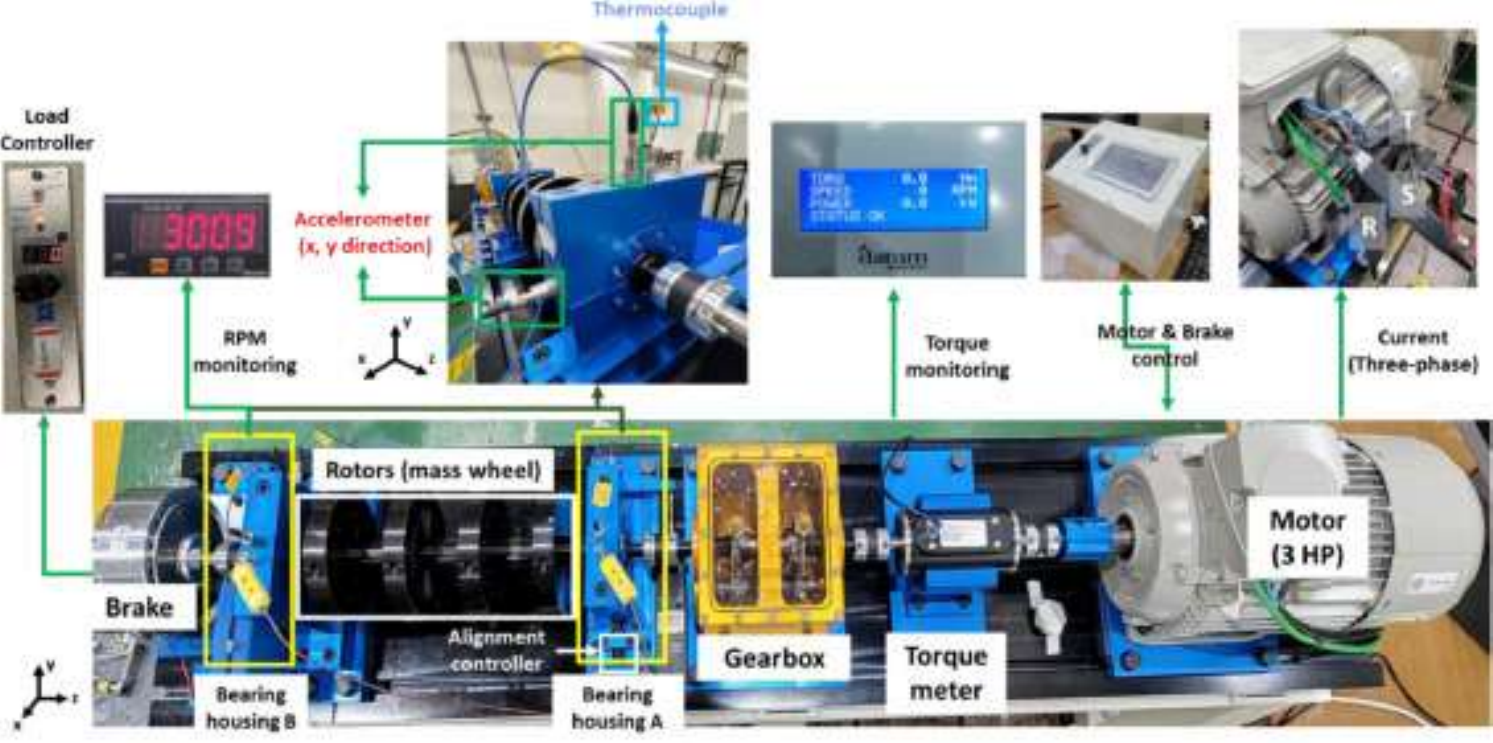}
        \caption{Rotating machinery platform of KAIST \cite{jung2023vibration} }
        \label{fig:korea}
\end{figure}

To further validate the diagnostic effectiveness of the proposed ISGFAN, the rotating machine dataset from the Korea Advanced Institute of Science and Technology (KAIST) was used in this experiment \cite{jung2023vibration}, and is available at \url{https://data.mendeley.com/datasets/ztmf3m7h5x/6}. The KAIST test platform is illustrated in Figure \ref{fig:korea}. Vibration data were measured using four accelerometers (PCB352C34) positioned at two bearing housings (A and B) in both the x and y directions. The vibration data were sampled at a frequency of 25.6 kHz. The states of the rotating machinery include normal operation, inner race fault, outer race fault, shaft misalignment, and rotor imbalance, as well as different severity levels of the same fault state. The torque loads for each condition are 0, 2, and 4 Nm. In this experiment, 2560 sampling points were used as time segments (samples), yielding 600 samples for each class. The vibration data from the KAIST dataset followed the same fault categories across all load conditions (0 Nm, 2 Nm, and 4 Nm), as detailed in Table \ref{tab:vibration_data_KAIST}.

\begin{table*}[!ht]
\small
\footnotesize
\caption{Comparison of models' performance: Transfer accuracy achieved by each method on the KAIST dataset across various tasks.}
\setlength{\tabcolsep}{3pt}  
\renewcommand{\arraystretch}{0.75}
\begin{tabular*}{\textwidth}{@{\extracolsep{\fill}} l >{\arraybackslash}p{2cm} | *{7}{c} }  
\toprule
\textbf{Noise type} & \multicolumn{1}{l}{\textbf{Model}} & \textbf{0-2} & \textbf{2-0} & \textbf{0-4} & \textbf{4-0} & \textbf{2-4} & \textbf{4-2} & \textbf{Average} \\
\midrule
\multirow{7}{*}{Gaussian (-8 dB)}
& AMARSN   & $73.32\%$  & $77.85\%$ & $69.05\%$  & $70.82\%$  & $85.25\%$  & $78.22\%$  & $75.75\%$ \\
& NORN     & $71.95\%$  & $71.15\%$ & $67.64\%$  & $70.26\%$  & $76.92\%$  & $73.42\%$  & $71.89\%$ \\
& AGFCN    & $72.06\%$  & $69.26\%$ & $66.95\%$  & $68.62\%$  & $81.35\%$  & $77.42\%$  & $72.61\%$ \\
& ITFS     & $81.02\%$  & $80.42\%$ & $78.45\%$  & $82.86\%$  & $89.55\%$  & $83.25\%$  & $82.59\%$ \\
& DSAN     & $82.35\%$  & $79.89\%$ & $77.12\%$  & $84.21\%$  & $88.92\%$  & $84.73\%$  & $82.87\%$ \\
& PEOTL    & $75.75\%$  & $80.15\%$ & $73.15\%$  & $73.65\%$  & $87.35\%$  & $80.42\%$  & $78.41\%$ \\
& \textbf{ISGFAN}   & $\textbf{84.62\%}$  & $\textbf{85.62\%}$ & $\textbf{80.82\%}$  & $\textbf{82.32\%}$  & $\textbf{94.13\%}$  & $\textbf{88.98\%}$  & $\textbf{86.08\%}$ \\
\cmidrule(l){2-9} 
\multirow{7}{*}{Laplacian (-8 dB)}
& AMARSN   & $71.75\%$  & $73.38\%$ & $68.91\%$  & $70.56\%$  & $81.16\%$  & $76.88\%$  & $73.77\%$ \\
& NORN     & $71.42\%$  & $71.23\%$ & $65.89\%$  & $73.34\%$  & $77.49\%$  & $72.17\%$  & $71.92\%$ \\
& AGFCN    & $70.31\%$  & $73.42\%$ & $66.59\%$  & $70.51\%$  & $80.64\%$  & $76.51\%$  & $73.00\%$ \\
& ITFS     & $80.06\%$  & $79.05\%$ & $77.38\%$  & $80.78\%$  & $87.23\%$  & $82.19\%$  & $81.12\%$ \\
& DSAN     & $79.23\%$  & $77.86\%$ & $74.94\%$  & $81.15\%$  & $85.08\%$  & $79.56\%$  & $79.64\%$ \\
& PEOTL    & $74.98\%$  & $79.13\%$ & $71.53\%$  & $73.10\%$  & $85.74\%$  & $78.82\%$  & $77.22\%$ \\
& \textbf{ISGFAN}   & $\textbf{84.16\%}$  & $\textbf{85.42\%}$ & $\textbf{80.68\%}$  & $\textbf{82.56\%}$  & $\textbf{90.45\%}$  & $\textbf{87.65\%}$  & $\textbf{85.15\%}$ \\
\cmidrule(l){2-9} 
\multirow{7}{*}{Mixed (-8 dB)}
& AMARSN   & $70.52\%$  & $72.78\%$ & $66.07\%$  & $68.02\%$  & $80.58\%$  & $74.66\%$  & $72.11\%$ \\
& NORN     & $69.62\%$  & $68.52\%$ & $61.09\%$  & $67.77\%$  & $75.31\%$  & $72.60\%$  & $69.15\%$ \\
& AGFCN    & $70.99\%$  & $67.65\%$ & $65.28\%$  & $68.25\%$  & $79.73\%$  & $74.06\%$  & $70.99\%$ \\
& ITFS     & $78.90\%$  & $77.69\%$ & $76.71\%$  & $80.95\%$  & $86.66\%$  & $82.12\%$  & $80.51\%$ \\
& DSAN     & $78.15\%$  & $74.45\%$ & $72.83\%$  & $79.32\%$  & $83.25\%$  & $78.89\%$  & $77.82\%$ \\
& PEOTL    & $73.87\%$  & $78.66\%$ & $70.78\%$  & $72.39\%$  & $86.41\%$  & $77.39\%$  & $76.58\%$ \\
& \textbf{ISGFAN}   & $\textbf{84.29\%}$  & $\textbf{85.52\%}$ & $\textbf{80.33\%}$  & $\textbf{81.45\%}$  & $\textbf{90.77\%}$  & $\textbf{87.79\%}$  & $\textbf{85.03\%}$ \\
\cmidrule(l){2-9} 
\bottomrule
\end{tabular*}
\label{tab: comparison_KAIST}
\end{table*}

\subsubsection{Test Results and Analysis} 
\begin{figure}
    \centering
    \includegraphics[width=1\linewidth]{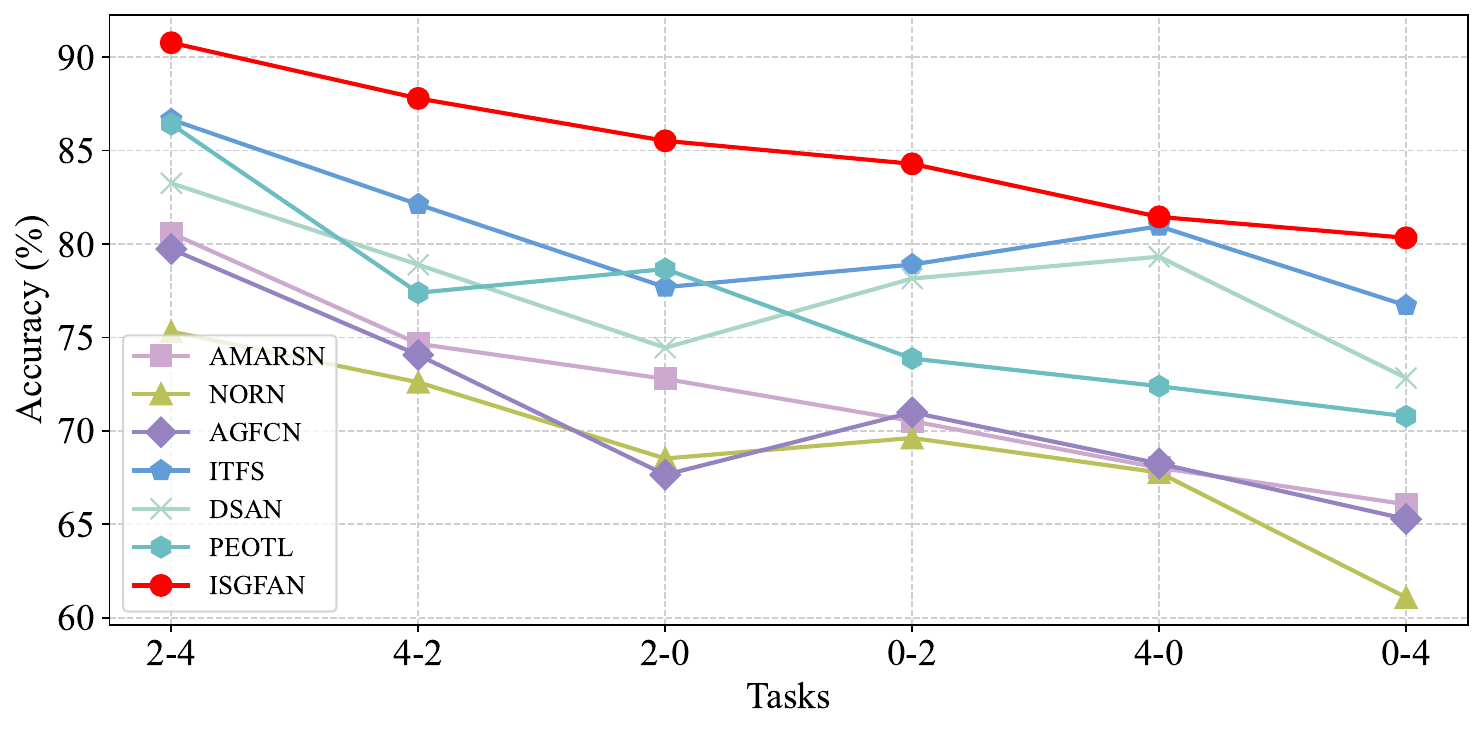}
    \caption{Performance comparison of each method under mixed noise conditions across transfer tasks on the KAIST dataset.}
    \label{fig:diagnosis_results}
\end{figure}

Table \ref{tab: comparison_KAIST} presents transfer accuracy results on KAIST datasets under different noise conditions. Under Gaussian noise, ISGFAN achieves the highest accuracy of 86.08\%, followed by DSAN at 82.87\%, ITFS at 82.59\%, PEOTL at 78.41\%, AMARSN at 75.75\%, AGFCN at 72.61\%, and NORN at 71.89\%. Laplacian noise, which creates sharper interference, reduces most models' performance, with ITFS and DSAN dropping to 81.12\% and 79.64\%, respectively, while ISGFAN maintains robust performance at 85.15\%. Under the most challenging mixed noise conditions, ISGFAN retains 85.03\% accuracy, significantly outperforming ITFS at 80.51\% and DSAN at 77.82\%. Figure \ref{fig:diagnosis_results} visually demonstrates these performance differences across transfer tasks. ISGFAN's superior generalization is evident in specific scenarios: achieving 80.33\% in the challenging 0-4 mixed noise task and 90.77\% in the easier 2-4 scenario, both substantially higher than comparison models.

Interestingly, some models show counterintuitive behavior. NORN performs better under Laplacian noise in the 4-0 scenario with 73.34\% than under Gaussian noise with 70.26\%, while AGFCN achieves slightly higher average accuracy under Laplacian noise at 73.00\% versus Gaussian noise at 72.61\%. These deviations from the expected trend suggest that model robustness depends on specific noise-transfer interactions rather than following a universal pattern. Overall, ISGFAN consistently demonstrates the most stable transfer performance across all noise conditions.

\subsection{Case Study III }
\subsubsection{Dataset Description}

\begin{figure}
    \centering
    \includegraphics[width=1\linewidth]{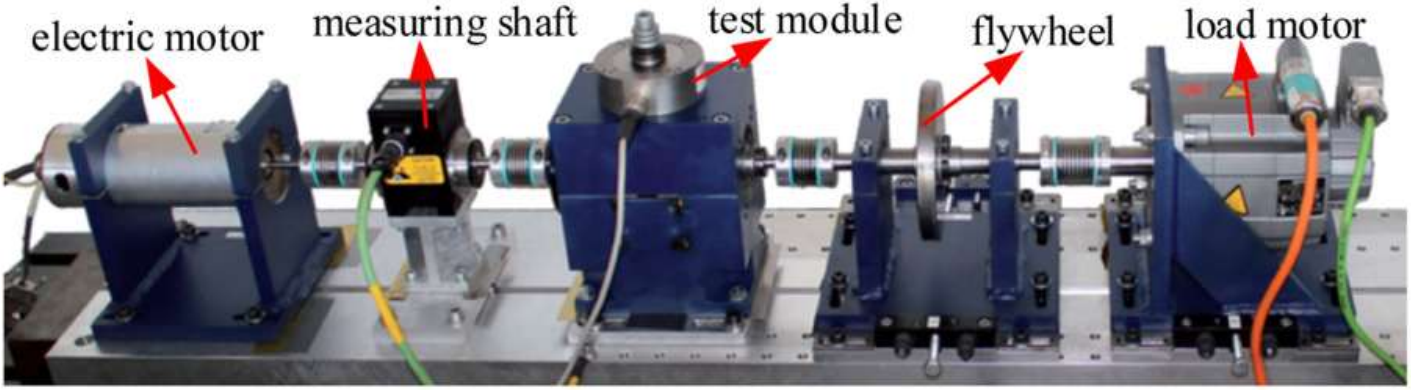}
    \caption{Rotating machinery fault testing and data acquisition platform of PU \citep{lessmeier2016condition}.}
    \label{fig:PU}
\end{figure}

The Paderborn University (PU) fault diagnosis dataset \citep{lessmeier2016condition} offers an extensive collection of bearing fault signal data, available at \url{https://mb.uni-paderborn.de/kat/forschung/kat-datacenter/bearing-datacenter}. As illustrated in Figure \ref{fig:PU}, the test rig comprises several integral components: an electric motor, a torque measurement shaft, a rolling bearing test module, a flywheel, and a load motor. The bearing faults are systematically categorized into two distinct types: artificially induced damages and naturally developed damages. Artificial damage mechanisms primarily include electrical discharge machining (EDM) for cracks, precision drilling for spalling, and electric engraving for pitting, whereas naturally degraded bearings are acquired through accelerated life testing platforms, with fault data sampled at 64 kHz. The PU dataset comprehensively captures complex damage conditions, including single damage, repetitive damage, and multiple damage scenarios. Table \ref{tab:PU_conditions} summarizes the fault data collected under four distinct operational conditions within the PU dataset framework. For the current experimental analysis, data from naturally degraded bearings were exclusively employed, with Table \ref{tab:fault_label_PU} delineating the specific fault categories utilized for testing purposes. Non-overlapping samples were created from the PU datasets, with each sample length modified to 1024. A total of 250 samples were allocated for each fault type. 

\begin{table}[!t]
    \small
\caption{Operational conditions for the PU datasets}
\setlength{\tabcolsep}{0pt}
\begin{tabular*}{\columnwidth}{@{\extracolsep{\fill}}ccccc}
\toprule
\begin{tabular}[c]{@{}c@{}}\textbf{Working}\\\textbf{condition}\end{tabular} & 
\begin{tabular}[c]{@{}c@{}}\textbf{Load}\\\textbf{(Nm)}\end{tabular} & 
\begin{tabular}[c]{@{}c@{}}\textbf{Radial force}\\\textbf{(N)}\end{tabular} & 
\begin{tabular}[c]{@{}c@{}}\textbf{Speed}\\\textbf{(rpm)}\end{tabular} & 
\begin{tabular}[c]{@{}c@{}}\textbf{Sample}\\\textbf{length}\end{tabular} \\
\midrule
0 & 0.7 & 1000 & 1500 &  1 $\times$ 1024 \\
1 & 0.7 & 1000 & 900 &  1 $\times$ 1024 \\
2 & 0.1 & 1000 & 1500 &  1 $\times$ 1024 \\
3 & 0.7 & 400 & 1500 &  1 $\times$ 1024 \\
\bottomrule
\end{tabular*}
\label{tab:PU_conditions}
\end{table}

\begin{table}[!t]  
    \small
    \caption{Fault categorization and sample quantities for the PU datasets}  
\setlength{\tabcolsep}{0pt}    
\begin{tabular*}{\columnwidth}{@{\extracolsep{\fill}}lcccc} 
\toprule  
\begin{tabular}[c]{@{}c@{}}\textbf{Bearing}\\\textbf{code}\end{tabular} & \begin{tabular}[c]{@{}c@{}}\textbf{Damage}\\\textbf{mode}\end{tabular} & \begin{tabular}[c]{@{}c@{}}\textbf{Damage}\\\textbf{position}\end{tabular} & \textbf{Combination} & \begin{tabular}[c]{@{}c@{}}\textbf{Sample}\\\textbf{quantity}\end{tabular} \\
\midrule  
KA04 & FP & OR & S & 250 \\
KA15 & PDI & OR & S & 250 \\
KA16 & FP & OR & R & 250 \\
KA22 & FP & OR & S & 250 \\
KA30 & PDI & OR & R & 250 \\
KB23 & FP & IR(+OR) & M & 250 \\
KB24 & FP & IR(+OR) & M & 250 \\
KB27 & PDI & OR + IR & M & 250 \\
KI04 & FP & IR & M & 250 \\
KI14 & FP & IR & M & 250 \\
KI16 & FP & IR & S & 250 \\
KI17 & FP & IR & R & 250 \\
KI18 & FP & IR & S & 250 \\
KI21 & FP & IR & S & 250 \\
\bottomrule  
\end{tabular*}  
\label{tab:fault_label_PU}  
\end{table}

\begin{table*}[t]
\small
\footnotesize
\caption{Comparison of performance: Transfer accuracy achieved by each method on the PU dataset across various tasks.}
\setlength{\tabcolsep}{3pt}  
\renewcommand{\arraystretch}{0.75}
\begin{tabular*}{\textwidth}{@{\extracolsep{\fill}} l >{\arraybackslash}p{2cm} | *{7}{c} }  
\toprule
\textbf{Noise type} & \multicolumn{1}{l}{\textbf{Model}} & \textbf{0-1} & \textbf{0-2} & \textbf{0-3} & \textbf{1-2} & \textbf{1-3} & \textbf{2-3} & \textbf{Average} \\
\midrule
\multirow{7}{*}{Gaussian (-8 dB)}
& AMARSN   & $48.67\%$  & $74.43\%$ & $59.11\%$  & $64.34\%$  & $27.55\%$  & $51.72\%$  & $54.30\%$  \\
& NORN     & $38.37\%$  & $67.75\%$ & $53.34\%$  & $54.16\%$  & $30.78\%$  & $53.53\%$  & $49.66\%$  \\
& AGFCN    & $44.82\%$  & $76.13\%$ & $62.53\%$  & $66.87\%$  & $36.90\%$  & $56.34\%$  & $57.27\%$  \\
& ITFS     & $68.25\%$  & $89.63\%$ & $78.43\%$  & $86.39\%$  & $64.76\%$  & $69.63\%$  & $76.18\%$  \\
& DSAN     & $67.85\%$  & $89.87\%$ & $77.96\%$  & $86.75\%$  & $64.32\%$  & $70.01\%$  & $76.13\%$   \\
& PEOTL    & $59.74\%$  & $81.48\%$ & $71.65\%$  & $79.32\%$  & $57.63\%$  & $65.66\%$  & $69.25\%$  \\
& \textbf{ISGFAN}   & $\textbf{69.68\%}$  & $\textbf{93.58\%}$ & $\textbf{83.64\%}$  & $\textbf{87.23\%}$  & $\textbf{71.74\%}$  & $\textbf{74.44\%}$  & $\textbf{80.05\%}$  \\
\cmidrule(l){2-9}  
\multirow{7}{*}{Laplacian (-8 dB)}
& AMARSN   & $47.52\%$  & $73.17\%$ & $57.85\%$  & $63.08\%$  & $31.32\%$  & $51.61\%$  & $54.09\%$  \\
& NORN     & $37.21\%$  & $66.51\%$ & $52.12\%$  & $52.93\%$  & $29.45\%$  & $52.28\%$  & $48.42\%$  \\
& AGFCN    & $44.59\%$  & $75.89\%$ & $62.31\%$  & $66.64\%$  & $36.67\%$  & $56.12\%$  & $57.04\%$  \\
& ITFS     & $67.03\%$  & $88.39\%$ & $77.21\%$  & $85.17\%$  & $63.51\%$  & $68.41\%$  & $74.95\%$  \\
& DSAN     & $63.78\%$  & $86.71\%$ & $79.68\%$  & $83.89\%$  & $61.87\%$  & $67.96\%$  & $73.98\%$ \\
& PEOTL    & $58.51\%$  & $80.24\%$ & $70.42\%$  & $78.08\%$  & $56.39\%$  & $64.43\%$  & $68.01\%$  \\
& \textbf{ISGFAN}   & $\textbf{68.45\%}$  & $\textbf{92.35\%}$ & $\textbf{82.41\%}$  & $\textbf{86.00\%}$  & $\textbf{71.91\%}$  & $\textbf{74.21\%}$  & $\textbf{79.22\%}$  \\
\cmidrule(l){2-9}  
\multirow{7}{*}{Mixed (-8 dB)}
& AMARSN   & $46.37\%$  & $71.91\%$ & $56.59\%$  & $56.82\%$  & $30.11\%$  & $50.50\%$  & $52.05\%$  \\
& NORN     & $37.05\%$  & $64.27\%$ & $49.90\%$  & $52.70\%$  & $28.12\%$  & $51.03\%$  & $47.18\%$  \\
& AGFCN    & $42.36\%$  & $73.65\%$ & $60.09\%$  & $64.41\%$  & $34.44\%$  & $53.90\%$  & $54.81\%$  \\
& ITFS     & $65.81\%$  & $87.15\%$ & $75.99\%$  & $83.95\%$  & $62.26\%$  & $67.19\%$  & $73.73\%$  \\
& DSAN     & $62.56\%$  & $87.48\%$ & $73.42\%$  & $81.21\%$  & $58.59\%$  & $66.85\%$  & $71.68\%$   \\
& PEOTL    & $57.28\%$  & $79.00\%$ & $69.19\%$  & $76.84\%$  & $55.15\%$  & $63.20\%$  & $66.78\%$  \\
& \textbf{ISGFAN}   & $\textbf{67.22\%}$  & $\textbf{92.12\%}$ & $\textbf{81.18\%}$  & $\textbf{87.77\%}$  & $\textbf{70.08\%}$  & $\textbf{71.98\%}$  & $\textbf{78.39\%}$  \\
\cmidrule(l){2-9} 
\bottomrule
\end{tabular*}
\label{tab: diagnostic_accuracy_PU}
\end{table*}

\subsubsection{ Test Results and Analysis}
\begin{figure}
    \centering
    \includegraphics[width=1\linewidth]{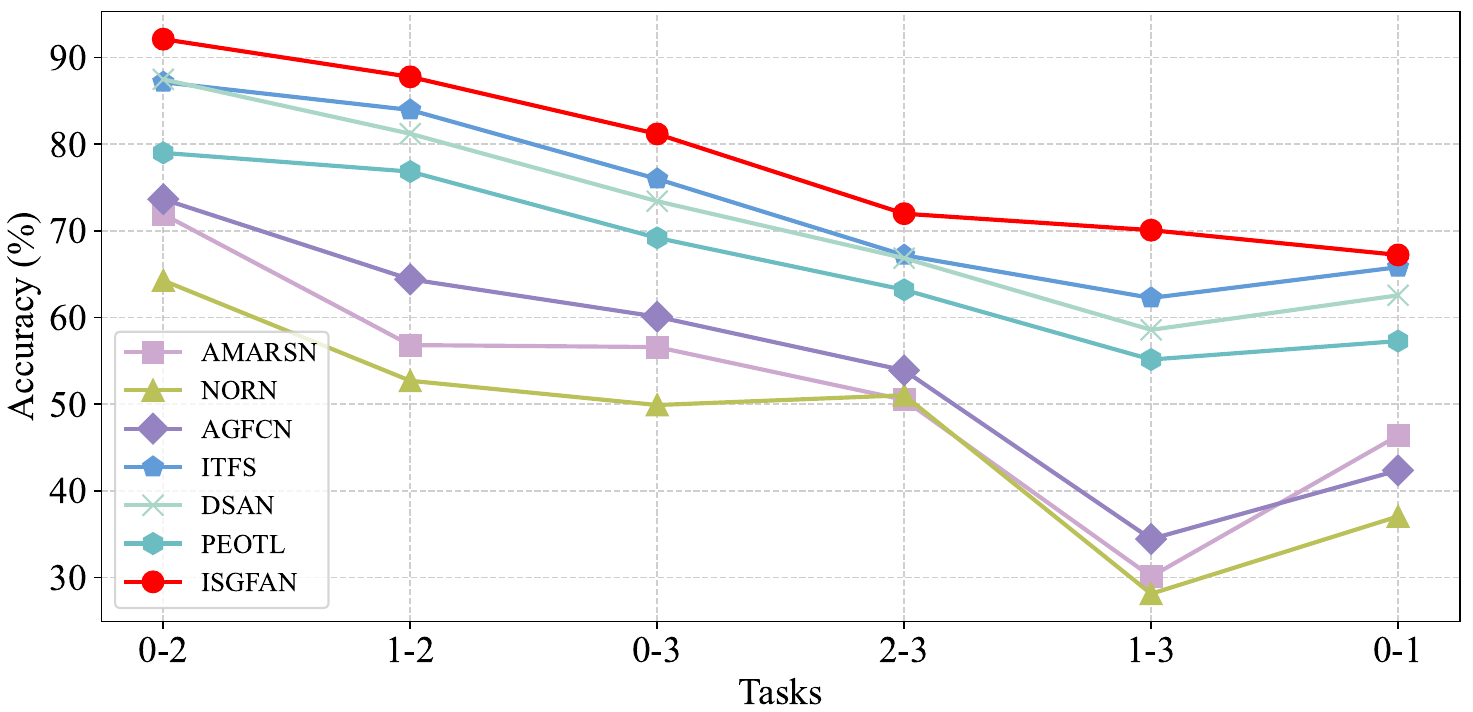}
    \caption{Performance comparison of each method under mixed noise conditions across transfer tasks on the PU dataset.}
    \label{fig:lines}
\end{figure}

Table \ref{tab: diagnostic_accuracy_PU} summarizes the transfer performance of each method on the PU dataset. Compared to the CWRU and KAIST datasets, the PU dataset exhibits a greater domain gap, and the feature differences between categories are less pronounced due to the specific types of damage. Consequently, knowledge transfer is more challenging under the influence of noise. The ISGFAN model demonstrates the strongest robustness across all types of noise and transfer tasks, achieving an average accuracy of 80.05\% under Gaussian noise, 79.22\% under Laplacian noise, and 78.39\% under mixed noise conditions. Among the baseline methods, ITFS exhibits competitive performance with average accuracies of 76.18\%, 74.95\%, and 73.73\%, respectively. All models consistently perform best in the 0-2 transfer task and worst in the 1-3 task, indicating significant variations in noise sensitivity across different transfer scenarios. In the 0-2 transfer task under mixed noise conditions, the accuracies achieved by AMARSN, NORN, AGFCN, ITFS, DSAN, PEOTL, and ISGFAN are 71.91\%, 64.27\%, 73.65\%, 87.15\%, 87.48\%, 79.00\%, and 92.12\%, respectively. Conversely, in the more challenging 1-3 transfer task, the corresponding accuracies decline substantially to 30.11\%, 28.12\%, 28.44\%, 62.26\%, 58.59\%, 55.15\%, and 70.08\%. As illustrated in Figure \ref{fig:lines}, when comparing the 0-2 and 1-3 transfer tasks, ISGFAN exhibits the smallest performance degradation and maintains stability. The anti-noise fault diagnosis models demonstrate performance comparable to transfer learning approaches when the domain gap is minimal; however, they experience more pronounced performance deterioration when confronting substantial domain discrepancies. Additionally, the conventional transfer models exhibit inferior noise resistance compared to ISGFAN. These experimental results validate the superior robustness of ISGFAN in scenarios characterized by concurrent noise interference and domain shift challenges.

\section{Conclusion} \label{sec:Conclusion}

Current fault diagnosis research for rotating machinery has not addressed the simultaneous challenges of noise interference and domain shift. To bridge this gap, we propose ISGFAN, which employs an information separation framework that combines adversarial learning with improved orthogonal constraints to produce domain-invariant representations, mitigating the effects of noise and domain-specific influences on classification and transfer processes. ISGFAN incorporates a global-focal domain-adversarial module that aligns both conditional and marginal distributions. The focal component uses SDC and SAA with pseudo-labeling to address class-specific transfer challenges in unsupervised scenarios, while the global component employs a domain discriminator for overall distribution alignment. Experiments on CWRU, KAIST, and PU datasets demonstrated ISGFAN's superior performance across varying noise conditions and operational scenarios, outperforming all baselines. Ablation studies and visualizations confirmed each component's effectiveness. However, limitations remain: the effectiveness of focal adaptation depends on pseudo-label quality, and SAA requires complex parameter tuning, necessitating extensive experimentation. Future work will focus on optimizing ISGFAN's architecture and reducing training complexity to develop a more streamlined, effective, and noise-robust transfer fault diagnosis model.

\section*{CRediT authorship contribution statement}
\textbf{Junyu Ren}: Writing – original draft, Methodology, Conceptualization. 
 \textbf{Wensheng Gan}: Writing – review \& editing, Supervision. 
 \textbf{Guangyu Zhang}: Validation. 
 \textbf{Wei Zhong}: Visualization.
 \textbf{Philip S. Yu}: Review and editing.

\section*{Declaration of competing interest}
The authors declare that they have no known competing financial interests or personal relationships that could have appeared to influence the work reported in this paper.

\section*{Acknowledgment}
This research was supported in part by National Natural Science Foundation of China (No. 62272196), Guangzhou Basic and Applied Basic Research Foundation (No. 2024A04J9971).

\section*{Data availability}
Data and code are available at \url{https://github.com/JYREN-Source/ISGFAN}

\bibliographystyle{model1-num-names}
\bibliography{main}

\end{document}